\documentclass[runningheads]{llncs}
\usepackage{graphicx}
\usepackage{tikz}
\usepackage{comment}
\usepackage{amsmath,amssymb} % define this before the line numbering.
\usepackage{color}

\usepackage[width=122mm,left=12mm,paperwidth=146mm,height=193mm,top=12mm,paperheight=217mm]{geometry}

\usepackage{xcolor}
\definecolor{darkgreen}{rgb}{0, 0.5, 0}
\definecolor{myblue}{RGB}{47, 114, 193}
\definecolor{brickred}{rgb}{0.8, 0.25, 0.33}
\definecolor{brandeisblue}{rgb}{0.0, 0.44, 1.0}
\definecolor{blueish}{rgb}{0.0, 0.3, .6}
\definecolor{pink}{rgb}{1, 0, 1}
\usepackage[pagebackref=true,breaklinks=true,colorlinks,citecolor=blueish,linkcolor=brickred,bookmarks=false]{hyperref}
\usepackage{booktabs}
\usepackage{multirow}
\usepackage{pifont}% http://ctan.org/pkg/pifont

\def\eg{\emph{e.g.}, } 
\def\ie{\emph{i.e.}, }
 
\def\etc{\emph{etc.} } \def\vs{\emph{vs.} }

\usepackage{enumitem}
\usepackage{algorithm}
\usepackage{algorithmic}
\usepackage{verbatim}
\usepackage{listings}
\usepackage{wrapfig}
\definecolor{codeblue}{rgb}{0.25,0.5,0.5}
\definecolor{codekw}{rgb}{0.85, 0.18, 0.50}
\usepackage{subcaption}
\usepackage{wrapfig}

\usepackage{mathtools}  
\usepackage{xfrac}  
\usepackage{booktabs}
\usepackage{cite}
\begin{document}
% \linenumbers
\pagestyle{headings}
\mainmatter
\def\ECCVSubNumber{3370}  % Insert your submission number here
% \def\ECCVSubNumber{xxxx}

% \title{Author Guidelines for ECCV Submission} % Replace with your title
% \title{Stand-Alone Multi-Axis Attention \\ Improves Vision Models}
% \title{Multi-Axis Attention Improves Vision Models}
% \title{Stand-Alone Multi-Axis Attention for Vision}
% \title{MAT: Multi-Axis Attention for Vision}
\title{MaxViT: Multi-Axis Vision Transformer}
% \title{Multi-Axis Vision Transformer}

\begin{comment}
\titlerunning{ECCV-22 submission ID \ECCVSubNumber} 
\authorrunning{ECCV-22 submission ID \ECCVSubNumber} 
\author{Anonymous ECCV submission}
\institute{Paper ID \ECCVSubNumber}
\end{comment}
%******************

% \begin{comment}
\titlerunning{MaxViT: Multi-Axis Vision Transformer}
% \author{Zhengzhong Tu\inst{1,2}\orcidID{0000-1111-2222-3333} \and
\author{Zhengzhong Tu\inst{1,2} \and
Hossein Talebi\inst{1} \and
Han Zhang\inst{1} \and
Feng Yang\inst{1} \and \\
Peyman Milanfar\inst{1} \and
Alan Bovik\inst{2} \and
Yinxiao Li\inst{1}}
\authorrunning{Z. Tu et al.}
\institute{Google Research \and
University of Texas at Austin
}
% \end{comment}
%******************
\maketitle

% \vspace{-6mm}
\begin{abstract}
% The abstract should summarize the contents of the paper. LNCS guidelines
% indicate it should be at least 70 and at most 150 words. It should be set in 9-point
% font size and should be inset 1.0~cm from the right and left margins.
% \dots
Transformers have recently gained significant attention in the computer vision community.
%
% Despite having larger model capacity than standard ConvNets, Transformers have been observed to underperform in generalization on ImageNet classification.
%
However, the lack of scalability of self-attention mechanisms with respect to image size has limited their wide adoption in state-of-the-art vision backbones.
In this paper we introduce an efficient and scalable attention model we call multi-axis attention, which consists of two aspects: blocked local and dilated global attention. These design choices allow global-local spatial interactions on arbitrary input resolutions with only linear complexity.
We also present a new architectural element by effectively blending our proposed attention model with convolutions, and accordingly propose a simple hierarchical vision backbone, dubbed MaxViT, by simply repeating the basic building block over multiple stages.
Notably, MaxViT is able to ``see'' globally throughout the entire network, even in earlier, high-resolution stages.
We demonstrate the effectiveness of our model on a broad spectrum of vision tasks.
%
%
% Here we present yet another hierarchical vision backbone ConvMaxViT, that effectively blends Convolution (Conv) with our proposed multi-axis self-attention/Transformer (MaxViT).
% MaxViT can be used a stand-alone sparse attention that enjoys a global-local perception on arbitrary input resolution with only a linear complexity to image size, which enables ConvMaxViT to have global interaction even in early, high-resolution stages. 
% Using a simple identical design of depthwise Conv with MaxViT throughout the entire network is surprisingly effective in improving generalization, capacity, and efficiency on image recognition.
On image classification, MaxViT achieves state-of-the-art performance under various settings: without extra data, MaxViT attains 86.5\% ImageNet-1K top-1 accuracy; with ImageNet-21K pre-training, our model achieves 88.7\% top-1 accuracy.
% (3) when pre-trained on larger-scale JFT-300M, MaxViT achieves xx.x\% top-1 accuracy. 
%
For downstream tasks, MaxViT as a backbone delivers favorable performance on object detection as well as visual aesthetic assessment.
We also show that our proposed model expresses strong generative modeling capability on ImageNet, demonstrating the superior potential of MaxViT blocks as a universal vision module.
%
% We will make the code and models publicly available.
The source code and trained models will be available at \url{https://github.com/google-research/maxvit}.

\keywords{Transformer, Image classification, Multi-axis attention.}
\end{abstract}

\section{Introduction}
Convolutional Neural Networks (ConvNets) have been the dominant architectural design choice for computer vision~\cite{krizhevsky2012imagenet,he2016deep,szegedy2015going,szegedy2016rethinking} since AlexNet~\cite{krizhevsky2012imagenet}.
ConvNets continue to excel on numerous vision problems by going deeper~\cite{szegedy2015going}, wider~\cite{szegedy2016rethinking,szegedy2017inception}, adding dense connections~\cite{huang2017densely}, efficient separable convolutions~\cite{howard2017mobilenets,sandler2018mobilenetv2}, atrous convolutions~\cite{chen2017deeplab}, using encoder-decoder frameworks~\cite{ronneberger2015u}, and even introducing modern micro-design components~\cite{liu2022convnet}.
Meanwhile, as inspired by the evolution of self-attention models like Transformers~\cite{vaswani2017attention} in natural language processing~\cite{devlin2018bert,yang2019xlnet,lan2019albert,raffel2019exploring}, numerous researchers have started to introduce attention mechanisms into vision~\cite{wang2018non,carion2020end}. 
The Vision Transformer (ViT)~\cite{dosovitskiy2020image} is perhaps the first fully Transformer-based architecture for vision, whereby image patches are simply regarded as sequences of words and a transformer encoder is applied on these visual tokens. When pre-trained on large-scale datasets~\cite{sun2017revisiting}, ViT can achieve compelling results on image recognition.

However, it has been observed that without extensive pre-training~\cite{dosovitskiy2020image,touvron2021training} ViT underperforms on image recognition. This is due to the strong model capacity of Transformers, that is imbued with less inductive bias, which leads to overfitting.
To properly regularize the model capacity and improve its scalability, numerous subsequent efforts have studied sparse Transformer models tailored for vision tasks such as local attention~\cite{liu2021swin,yang2021focal,li2021localvit,chu2021twins}. These methods typically re-introduce hierarchical architectures to compensate for the loss of non-locality.
The Swin Transformer~\cite{liu2021swin} is one such successful attempt to modify Transformers by applying self-attention on shifted non-overlapping windows. For the first time, this approach outperformed ConvNets on the ImageNet benchmark with a pure vision Transformer.
Despite having more flexibility and generalizability than the full attention used in ViT, window-based attention has been observed to have limited model capacity due to the loss of non-locality, and henceforth scales unfavorably on larger data regimes such as ImageNet-21K and JFT~\cite{dai2021coatnet}.
However, acquiring global interactions via full-attention at early or high-resolution stages in a hierarchical network is computationally heavy, as the attention operator requires quadratic complexity.
How to efficiently incorporate global and local interactions to balance the model capacity and generalizability under a computation budget still remains challenging.
% Balancing the trade-off between model capacity and generalization ability given a computation budget is still very challenging.
%

\begin{figure}[!tb]
 \centering
 \begin{subfigure}[b]{0.49\textwidth}
     \centering
     \includegraphics[width=\textwidth]{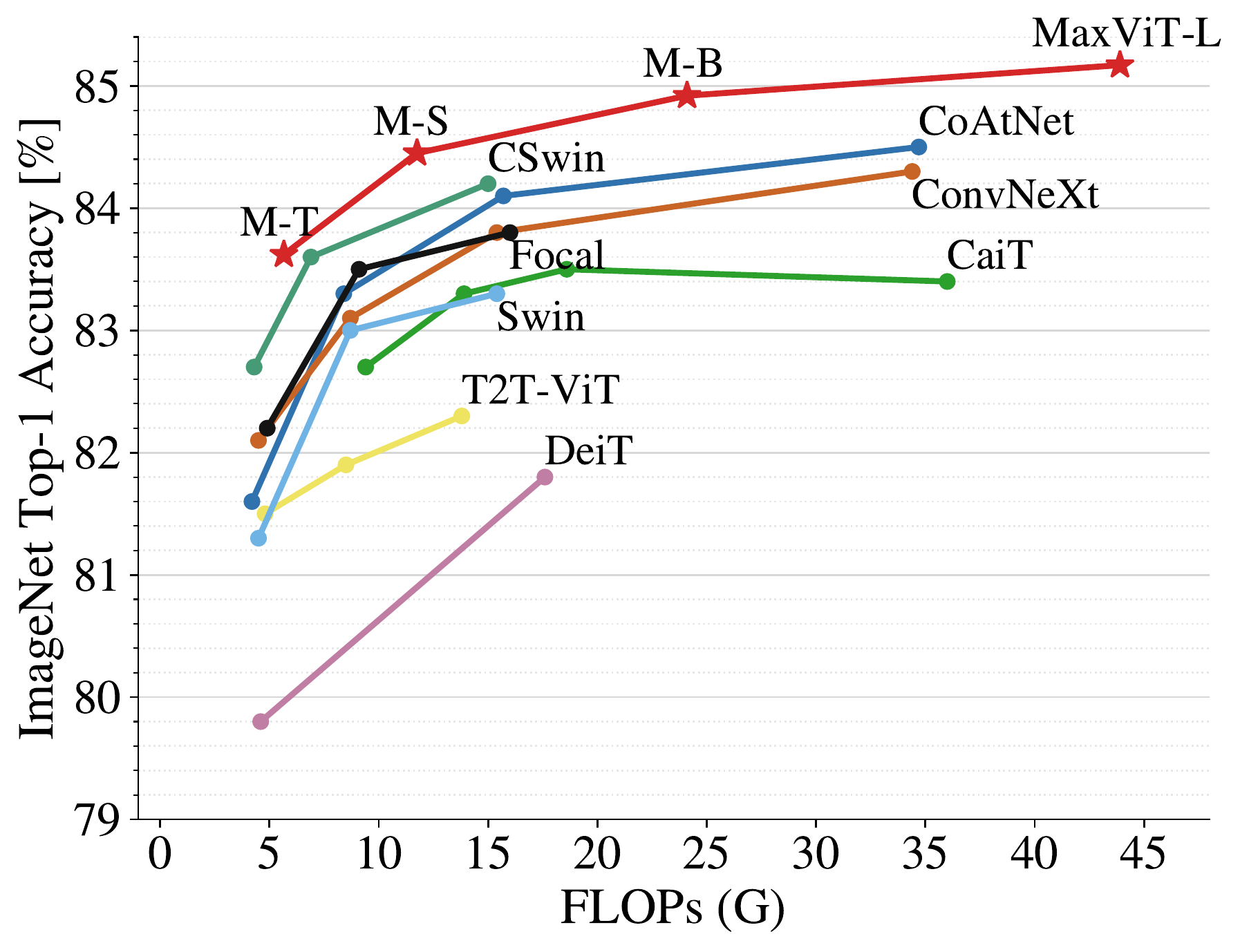}
     \caption{Accuracy \vs FLOPs performance scaling curve under ImageNet-1K training setting at input resolution 224$\times$224.}
     \label{fig:imagenet-flops}
 \end{subfigure}
 \hfill
 \begin{subfigure}[b]{0.49\textwidth}
     \centering
     \includegraphics[width=\textwidth]{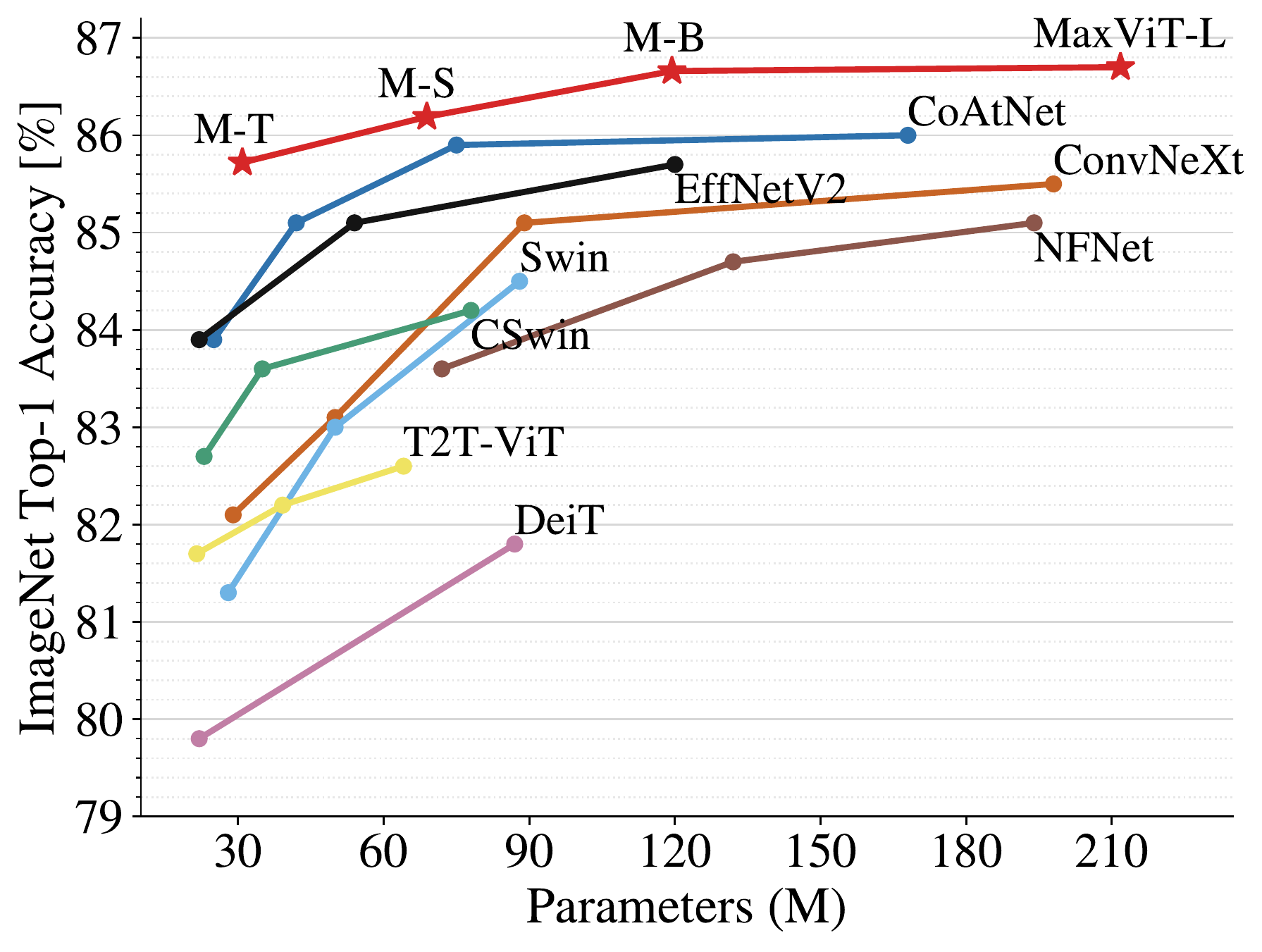}
     \caption{Accuracy \vs Parameters scaling curve under ImageNet-1K fine-tuning setting allowing for higher sizes (384/512).}
     \label{fig:three sin x}
 \end{subfigure}
\caption{\textbf{Performance comparison of MaxViT with state-of-the-art vision Transformers on ImageNet-1K.} Our model shows superior performance in terms of both accuracy \vs computation and accuracy \vs parameters tradeoff.}
\label{fig:imagenet-1k-main}
\end{figure}

In this paper, we present a new type of Transformer module, called multi-axis self-attention (Max-SA), that capably serves as a basic architecture component which can perform both local and global spatial interactions in a single block.
Compared to full self-attention, Max-SA enjoys greater flexibility and efficiency, \ie naturally adaptive to different input lengths with linear complexity; in contrast to (shifted) window/local attention, Max-SA allows for stronger model capacity by proposing a global receptive field.
Moreover, with merely linear complexity, Max-SA can be used as a general stand-alone attention module in any layer of a network, even in earlier, high-resolution stages.

To demonstrate its effectiveness and universality, we further design a simple but effective vision backbone called \textbf{M}ulti-\textbf{ax}is \textbf{Vi}sion \textbf{T}ransformer (\textbf{MaxViT}) by hierarchically stacking repeated blocks composed of Max-SA and convolutions.
While our proposed model belongs to the category of hybrid vision Transformers, MaxViT distinguishes from previous approaches~\cite{xiao2021early,dai2021coatnet} in that we strive for simplicity, by designing a basic block unifying convolution, local, and global attention, then simply repeating it.
Our experiments shows that the MaxViT significantly improves upon state-of-the-art (SOTA) performance under all data regimes for a broad range of visual tasks including classification, object detection and segmentation, image aesthetics assessment, and image generation.
Specifically, as Figure~\ref{fig:imagenet-1k-main} shows, MaxViT outperforms all recent Transformer-based models in regards to both accuracy \vs FLOPs and accuracy \vs parameter curves. Our contributions are:

% \vspace{-2mm}
\begin{itemize}[leftmargin=*]
\itemsep0em 
    \item A generic strong Transformer backbone, \textbf{MaxViT}, that can capture both local and global spatial interactions throughout every stage of the network.
    % \vspace{-0.6mm}
    \item A novel stand-alone multi-axis attention module composed of blocked local and dilated global attention, enjoying global perception in linear complexity.
    % \vspace{-0.6mm}
    \item We demonstrate large amounts of design choices including number of layers, layouts, the use of MBConv, \etc with extensive ablation studies, that eventually converge towards our final modular design, the MaxViT-Block.
    \item Our extensive experiments show that MaxViT achieves SOTA results under various data regimes for a broad range of tasks including image classification, object detection, image aesthetic assessment, and image generation.

\end{itemize}

% \textcolor{red}{(Introducing Transformers in Vision)} 

% \textcolor{red}{The problem of ViT/Swin. Then introduce Hybrid models. Still, the problem of using global models on high-res images/features exist.}

% \textcolor{red}{We explore sparse attention models.}

\section{Related work}

\textbf{Convolutional networks.}
Since AlexNet~\cite{krizhevsky2012imagenet}, convolutional neural networks (ConvNets) have been used as \textit{de facto} solutions to almost all vision tasks~\cite{he2016deep,chen2020proxiqa,huang2017densely,zhao2022tracking,chen2022learning,li2021comisr,whang2022deblurring,talebi2021learning,wang2021rich} before the ``Roaring 20s''~\cite{liu2022convnet}. Phenomenal architectural improvements have been made in the past decade: residual~\cite{he2016deep} and dense connections~\cite{huang2017densely}, fully-convolutional networks~\cite{long2015fully}, encoder-decoder schemes~\cite{ronneberger2015u}, feature pyramids~\cite{Lin2017FeaturePN}, increased depths and widths~\cite{szegedy2015going}, spatial- and channel-wise attention models~\cite{hu2018squeeze,woo2018cbam}, non-local interactions~\cite{wang2018non}, to name a few. A remarkable recent work ConvNeXt~\cite{liu2022convnet} has re-introduced core designs of vision Transformers and shown that a `modernized' pure ConvNet can achieve  performance comparable to Transformers on broad vision tasks.

\noindent\textbf{Transformers in vision.}
Transformers were originally proposed for natural language processing~\cite{vaswani2017attention}. The debut of the Vision Transformer (ViT)~\cite{dosovitskiy2020image} in 2020 showed that pure Transformer-based architectures are also effective solutions for vision problems. The elegantly novel view of ViT that treats image patches as visual words has stimulated explosive research interest in visual Transformers. To account for locality and 2D nature of images, the Swin Transformer aggregates attention in shifted windows in a hierarchical architecture~\cite{liu2021swin}. More recent works have been focused on improving model and data efficiency, including sparse attention~\cite{dong2021cswin,yang2021focal,rao2021dynamicvit,wang2020axial, xu2022cobevt, arnab2021vivit}, improved locality~\cite{yuan2021tokens,han2021transformer}, pyramidal designs~\cite{wang2021pyramid,fan2021multiscale, xu2022v2x},
improved training strategies~\cite{touvron2021training,touvron2021going,zhou2021deepvit,bello2021revisiting}, \etc We refer readers to dedicated surveys~\cite{khan2021transformers,khan2021transformers} of vision Transformers for a comprehensive review.

\noindent\textbf{Hybrid models.}
Pure Transformer-based vision models have been observed to generalize poorly due to relatively less inductive bias~\cite{dosovitskiy2020image,dai2021coatnet,touvron2021training}. Vision Transformers also exhibit substandard optimizability~\cite{xiao2021early}. An intriguingly simple improvement is to adopt a hybrid design of Transformer and convolution layers such as using a few convolutions to replace the coarse patchify stem~\cite{xiao2021early,dai2021coatnet}. A broad range of works fall into this category, either explicitly hybridized~\cite{wu2021cvt,d2021convit,dai2021coatnet,xiao2021early,fan2021multiscale,xu2021co,bello2019attention} or in an implicit fashion~\cite{liu2021swin,chu2021twins}.

\noindent\textbf{Transformer for GANs.}
Transformers have also proven effective in generative adversarial networks (GANs)~\cite{goodfellow2014generative}. TransGAN~\cite{jiang2021transgan} built a pure Transformer GAN with a careful design of local attention and upsampling layers, demonstrating effectiveness on small scale datasets~\cite{krizhevsky2009learning,pmlr-v15-coates11a}. GANformer~\cite{hudson2021generative} explored efficient global attention mechanisms to improve on StyleGAN~\cite{karras2020analyzing} generator. HiT~\cite{zhao2021improved} presents an efficient Transformer generator based on local-global attention that can scale up to 1K high-resolution image generation.

\section{Method}

\begin{figure}[!t]
\centering
\includegraphics[width=0.96\linewidth]{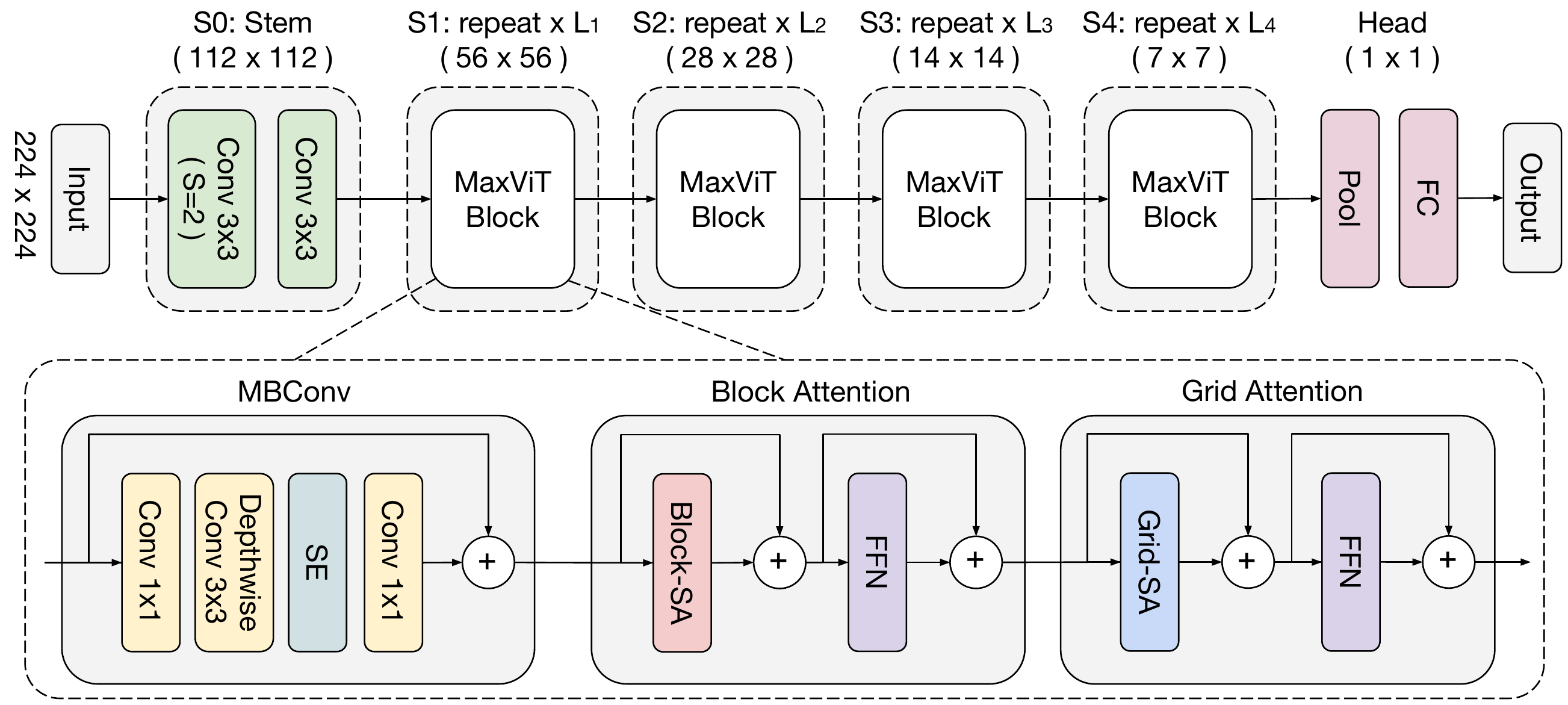}
\caption{\textbf{MaxViT architecture.} We follow a typical hierarchical design of ConvNet practices (e.g., ResNet) but instead build a new type of basic building block that unifies MBConv, block, and grid attention layers. Normalization and activation layers are omitted for simplicity.}
\label{fig:maxt-architecture}
%  \vspace{-3mm}
\end{figure}

% \textcolor{red}{describing the overall system by schematicing a general backbone similar to CoatNet, Swin or Uniformer.}
%
Inspired by the sparse approaches presented in~\cite{zhao2021improved,tu2022maxim}, we introduce a new type of attention module, dubbed blocked multi-axis self-attention (Max-SA), by decomposing the fully dense attention mechanisms into two sparse forms -- window attention and grid attention -- which reduces the quadratic complexity of vanilla attention to linear, without any loss of non-locality.
Our sequential design offers greater simplicity and flexibility, while performing even better than previous methods -- each individual module can be used either standalone or combined in any order (Tables \ref{tab:ablation-grid-attention}-\ref{tab:ablation-mbconv}), whereas parallel designs~\cite{zhao2021improved,tu2022maxim} offer no such benefits.
Because of the flexibility and scalability of Max-SA, we are able to build a novel vision backbone, which we call MaxViT, by simply stacking alternative layers of Max-SA with MBConv~\cite{howard2017mobilenets} in a hierarchical architecture, as shown in Figure~\ref{fig:maxt-architecture}.
MaxViT benefits from global and local receptive fields throughout the entire network, from shallow to deep stages, demonstrating superior performance in regards to both model capacity and generalization abilities.

\subsection{Attention}
Self-attention allows for spatial mixing of entire spatial (or sequence) locations while also benefiting from content-dependent weights based on normalized pairwise similarity. 
%
% Specifically, let $x_i,y_i\in\mathbb{R}^{D}$ be the input and output at position $i$, point-wise self-attention function is defined as:
% \begin{equation}
% \label{eq:self-attention}
% y_i=\sum_{j\in\mathcal{G}}\frac{\exp{x^\top_{i}x_j}}{\sum_{k\in\mathcal{G}}\exp(x_i^\top x_k)}x_j,
% \end{equation}
% where $\mathcal{G}$ denotes the global spatial locations.
%
The standard self-attention defined in \cite{vaswani2017attention,dosovitskiy2020image} is location-unaware, \ie non-translation equivariant, an important inductive bias imbued in ConvNets.
Relative self-attention~\cite{liu2021swin,dai2021coatnet,shaw2018self,jiang2021transgan} has been proposed to improve on vanilla attention by introducing a relative learned bias added to the attention weights, which has been shown to consistently outperform original attention on many vision tasks~\cite{liu2021swin,dai2021coatnet,jiang2021transgan}.
In this work, we 
mainly adopt the pre-normalized relative self-attention defined in~\cite{dai2021coatnet} as the key operator in MaxViT.
%
%
% Specifically, let $x_i,y_i\in\mathbb{R}^{D}$ be the input and output at position $i$, pre-normalized relative self-attention is defined as:
% \begin{equation}
% \label{eq:relative-attention}
% y_i^\text{pre-norm}=\sum_{j\in\mathcal{G}}\frac{\exp{(x^\top_{i}x_j+w_{i-j})}}{\sum_{k\in\mathcal{G}}\exp(x_i^\top x_k+{w_{i-k}})}x_j,
% \end{equation}
% where the attention weights are co-decided by a learned static location-aware kernel $w_{i-j}$ and the input-adaptive attention $x^\top_{i}x_j$, enjoying both input-adaptivity, translation equivariance, and global interactions. 

\subsection{Multi-axis Attention}

Global interaction is one of the key advantages of self-attention as compared to local convolution.
However, directly applying attention along the entire space is computationally infeasible as the attention operator requires quadratic complexity.
To tackle this problem, we present a multi-axis approach to decompose the full-size attention into two sparse forms -- local and global -- by simply decomposing the spatial axes.
Let $X\in \mathbb{R}^{H\times W\times C}$ be an input feature map. Instead of applying attention on the flattened spatial dimension $HW$, we {block} the feature into a tensor of shape $(\frac{H}{P}\times\frac{W}{P},P\times P,C)$, representing partitioning into non-overlapping windows, each of size $P\times P$.
Applying self-attention on the local spatial dimension~\emph{i.e.}, $P\times P$, is equivalent to attending within a small window~\cite{liu2021swin}.
We will use this \textbf{block attention} to conduct local interactions.

\begin{figure}[!t]
\centering
\includegraphics[width=0.96\linewidth]{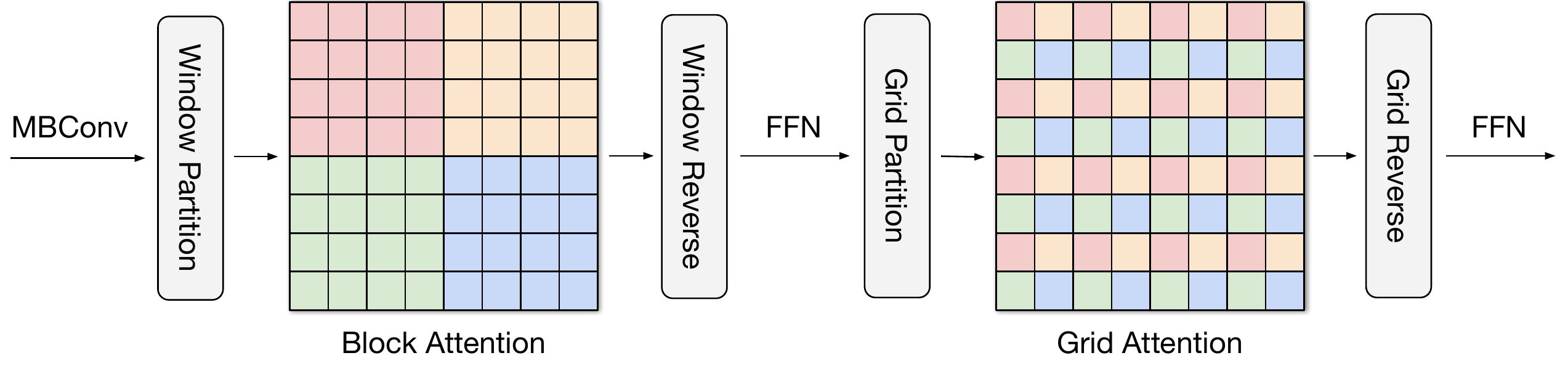}
\caption{\textbf{Multi-axis self-attention (Max-SA)} (best viewed in color). An illustration of the multi-axis approach for computing self-attention (window/grid size is 4$\times$4). The block-attention module performs self-attention within windows, while the grid-attention module attends globally to pixels in a sparse, uniform grid overlaid on the entire 2D space, with both having linear complexity against input size, as we use fixed attention footage. The same colors are spatially mixed by the self-attention operation.}
\label{fig:MaxViT-block}
%  \vspace{-3mm}
\end{figure}

Despite bypassing the notoriously heavy computation of full self-attention, local-attention models have been observed to underfit on huge-scale datasets~\cite{dai2021coatnet,dosovitskiy2020image}.
Inspired by block attention, we present a surprisingly simple but effective way to gain sparse global attention, which we call \textbf{grid attention}.
Instead of partitioning feature maps using fixed window size, we {grid} the tensor into the shape $(G\times G,\frac{H}{G}\times\frac{W}{G},C)$ using a fixed $G\times G$ uniform grid, resulting in windows having adaptive size $\frac{H}{G}\times\frac{W}{G}$.
Employing self-attention on the decomposed grid axis~\emph{i.e.}, $G\times G$, corresponds to dilated, global spatial mixing of tokens.
By using the same \textit{fixed} window and grid sizes (we use $P=G=7$ following Swin~\cite{liu2021swin}), we can fully balance the computation between local and global operations, both having only linear complexity with respect to spatial size or sequence length.
Note that our proposed Max-SA module can be a drop-in replacement of the Swin attention module~\cite{liu2021swin} with exactly the same number of parameters and FLOPs. Yet it enjoys \textit{global interaction} capability without requiring masking, padding, or cyclic-shifting, making it more implementation friendly, preferable to the shifted window scheme~\cite{liu2021swin}.
For instance, the multi-axis attention can be easily implemented with \texttt{einops}~\cite{rogozhnikov2022einops} without modifying the original attention operation (see Appendix).
It is worth mentioning that our proposed multi-axis attention (Max-SA) is fundamentally different from the axial-attention models~\cite{wang2020axial,ho2019axial}.
Please see Appendix for a detailed comparison.

% \begin{wrapfigure}{r}{0.5\textwidth}
% % \vspace{-15mm}
% \begin{minipage}{0.5\textwidth}
% \floatname{algorithm}{Algo.}
% \begin{algorithm}[H]
% \small
% \caption{\small Pseudocode of MaxViT Block}
% \label{alg:code}
% \begin{lstlisting}[language=python]
% # input: features (b, h, w, c). Assume h==w.
% # x/output: features (b, h, w, c).
% # p/g: block/grid size. Use 7 by default.

% # A self-attn function applied on the -2 axis
% def RelSelfAttn(x): return x # A placeholder

% # Window/grid partition function
% from einops import rearrange
% def block(x,p):
%   return rearrange(x,"b(hy)(wx)c->b(hw)(yx)c",
%     h=x.shape[1]//p,w=x.shape[2]//p,y=p,x=p)
    
% def unblock(x,g,p):
%   return rearrange(x,"b(hw)(yx)c->b(hy)(wx)c",
%     h=g,w=g,y=p,x=p) 

% x = MBConv(input) # MBConv layer

% x = block(x,p) # window partition
% x = RelSelfAttn(x) # Apply window-attention
% x = unblock(x,x.shape[1]//p,p) # reverse

% x = block(x,x.shape[1]//g) # grid partition
% x = swapaxes(x,-2,-3) # move grid-axis to -2
% x = RelSelfAttn(x) # Apply grid-attention
% x = swapaxes(x,-2,-3) # reverse swapaxes
% output = unblock(x,g,x.shape[1]//g) # reverse
% \end{lstlisting}
% \end{algorithm}
% \end{minipage}
% % \vspace{-5mm}
% \end{wrapfigure}

% \vspace{-5mm}
%
\noindent\textbf{MaxViT block.} We sequentially stack the two types of attentions to gain both local and global interactions in a single block, as shown in Figure~\ref{fig:MaxViT-block}.
Note that we also adopt typical designs in Transformers~\cite{dosovitskiy2020image,liu2021swin}, including LayerNorm~\cite{ba2016layer}, Feedforward networks (FFNs)~\cite{dosovitskiy2020image,liu2021swin}, and skip-connections.
We also add a MBConv block~\cite{howard2017mobilenets} with  squeeze-and-excitation (SE) module~\cite{hu2018squeeze} prior to the multi-axis attention, as we have observed that using MBConv together with attention further increases the generalization as well as the trainability of the network~\cite{xiao2021early}.
Using MBConv layers prior to attention offers another advantage, in that depthwise convolutions can be regarded as conditional position encoding (CPE)~\cite{chu2021conditional}, making our model free of explicit positional encoding layers.
%
% We demonstrate in Algo.~\ref{alg:code} an \texttt{einops}-style pseudocode of the MaxViT block.
%
Note that our proposed stand-alone multi-axis attention may be used together or in isolation for different purposes -- block attention for local interaction, and grid attention for global mixing.
These elements can be easily plugged into many vision architectures, especially on high-resolution tasks that can benefit by global interactions with affordable computation.

\subsection{Architecture Variants}
We designed a series of extremely simple architectural variants to explore the effectiveness of our proposed MaxViT block, as shown in Figure~\ref{fig:maxt-architecture}.
We use a hierarchical backbone similar to common ConvNet practices~\cite{he2016deep,liu2022convnet,dai2021coatnet,tan2021efficientnetv2} where the input is first downsampled using Conv3x3 layers in stem stage (S0).
The body of the network contains four stages (S1-S4), with each stage having half the resolution of the previous one with a doubled number of channels (hidden dimension).
In our network, we employ \textit{identical} MaxViT blocks throughout the entire backbone.
We apply downsampling in the Depthwise Conv3x3 layer of the first MBConv block in each stage.
The expansion and shrink rates for inverted bottleneck~\cite{howard2017mobilenets} and squeeze-excitation (SE)~\cite{hu2018squeeze} are 4 and 0.25 by default.
We set the attention head size to be 32 for all attention blocks. We scale up the model by increasing block numbers per stage $B$ and the channel dimension $C$.
We summarize the architectural configurations of the MaxViT variants in Table~\ref{tab:model-variants}.

\begin{table}[!t]
\centering
\footnotesize
\setlength{\tabcolsep}{1.2pt}
\renewcommand{\arraystretch}{1.0}
\caption{\normalsize \textbf{MaxViT architecture variants.} \texttt{B} and \texttt{C} denotes number of blocks and number of channels for each stage. We set each attention head to 32 for all attention layers. For MBConv, we always use expansion rate 4 and shrinkage rate 0.25 in SE~\cite{hu2018squeeze}, following~\cite{tan2019efficientnet,tan2021efficientnetv2,dai2021coatnet}. We use two Conv layers in the stem.}
\label{tab:model-variants}
\begin{tabular}{l|c|l|l|l|l|l}
{Stage} & {Size} & {MaxViT-T} & {MaxViT-S} & { MaxViT-B} & { MaxViT-L} & {MaxViT-XL} \\
\toprule
\texttt{S0}: Conv-stem & $\sfrac{1}{2}$ & \texttt{B=2  C=64} & \texttt{B=2 C=64} &  \texttt{B=2 \space  C=64} & \texttt{B=2 \space C=128}  & \texttt{B=2 \space C=192} \\
\texttt{S1}: MaxViT-Block & $\sfrac{1}{4}$ & \texttt{B=2  C=64} & \texttt{B=2 C=96} &  \texttt{B=2  \space  C=96} & \texttt{B=2 \space C=128}  & \texttt{B=2 \space C=192} \\
\texttt{S2}: MaxViT-Block & $\sfrac{1}{8}$ & \texttt{B=2  C=128} & \texttt{B=2 C=192} &  \texttt{B=6  \space  C=192} & \texttt{B=6 \space C=256}  & \texttt{B=6 \space C=384} \\
\texttt{S3}: MaxViT-Block & $\sfrac{1}{16}$ & \texttt{B=5  C=256} & \texttt{B=5 C=384} &  \texttt{B=14  C=384} & \texttt{B=14  C=512}  & \texttt{B=14  C=768} \\
\texttt{S4}: MaxViT-Block & $\sfrac{1}{32}$ & \texttt{B=2  C=512} & \texttt{B=2 C=768} &  \texttt{B=2  \space  C=768} & \texttt{B=2 \space C=1024}  & \texttt{B=2 \space C=1536} \\
% \bottomrule
\end{tabular}
\end{table}

% \vspace{-1mm}
% \begin{itemize}[leftmargin=*]
% \itemsep0em 
%     \item[\textcolor{blueish}{$\bullet$}] MaxViT-T: $B=[2, 2, 5, 2]$, $C_0,C_{1-4}=64, [64, 128, 256, 512]$, $P=G=7$
%     % \vspace{-0.3mm}
%     \item[\textcolor{blueish}{$\bullet$}] MaxViT-S: $B=[2, 2, 5, 2]$, $C_0,C_{1-4}=64, [96, 192, 384, 768]$, $P=G=7$
%     % \vspace{-0.3mm}
%     \item[\textcolor{blueish}{$\bullet$}] MaxViT-B: $B=[2, 6, 14, 2]$, $C_0,C_{1-4}=64, [96, 192, 384, 768]$, $P=G=7$
%     % \vspace{-0.3mm}
%     \item[\textcolor{blueish}{$\bullet$}] MaxViT-L: $B=[2, 6, 14, 2]$,
%     $C_0,C_{1-4}=128, [128, 256, 512, 1024]$, $P=G=7$
%     % \vspace{-0.3mm}
%     \item[\textcolor{blueish}{$\bullet$}] MaxViT-H: $B=[2, 6, 14, 2]$, $C_0,C_{1-4}=192, [192, 384, 768, 1536]$, $P=G=7$
% \end{itemize}

\section{Experiments}

We validated the efficacy of our proposed model on various vision tasks: ImageNet classification~\cite{krizhevsky2012imagenet}, image object detection and instance segmentation~\cite{lin2014microsoft}, image aesthetics/quality assessment~\cite{murray2012ava}, and unconditional image generation~\cite{goodfellow2014generative}. More experimental details can be found in the Appendix.

\begin{table*}[!t]
\centering
\footnotesize
\setlength{\tabcolsep}{3.6pt}
\renewcommand{\arraystretch}{0.9}
\caption{\normalsize \textbf{Performance comparison under ImageNet-1K setting.} Throughput is measured on a single V100 GPU with batch size 16, following~\cite{liu2021swin,tan2021efficientnetv2,liu2022convnet}.}
\label{tab:imagenet1k-comparison}
\begin{tabular}{c|l|ccccc}
% \hline
% \rowcolor[gray]{0.95}
 &
\multirow{2}{*}{Model} & \multirow{2}{*}{\begin{tabular}{@{}c@{}}Eval\\size\end{tabular}} & \multirow{2}{*}{Params} & \multirow{2}{*}{FLOPs} & \multirow{2}{*}{\begin{tabular}{@{}c@{}}Throughput \\(image/s)\\ \end{tabular}} &
\multirow{2}{*}{\begin{tabular}{@{}c@{}}IN-1K \\top-1 acc.\\ \end{tabular}} \\
&&&&\\
% \rowcolor[gray]{0.95}
\toprule
\multirow{10}{*}{ConvNets} & \textcolor{blueish}{$\bullet$}EffNet-B6~\cite{tan2019efficientnet} & 528  & 43M & 19.0G & 96.9  & 84.0 \\
& \textcolor{blueish}{$\bullet$}EffNet-B7~\cite{tan2019efficientnet} & 600  & 66M & 37.0G  &  55.1  & 84.3 \\
& \textcolor{blueish}{$\bullet$}RegNetY-16~\cite{radosavovic2020designing} & 224 & 84M & 16.0G & 334.7 & 82.9 \\
& \textcolor{blueish}{$\bullet$}NFNet-F0~\cite{brock2021high} & 256 & 72M & 12.4G  & 533.3 & 83.6 \\
& \textcolor{blueish}{$\bullet$}NFNet-F1~\cite{brock2021high} & 320 & 132M & 35.5G  & 228.5 &  84.7 \\
& \textcolor{blueish}{$\bullet$}EffNetV2-S~\cite{tan2021efficientnetv2} & 384 & 24M & 8.8G  & 666.6 &  83.9 \\
& \textcolor{blueish}{$\bullet$}EffNetV2-M~\cite{tan2021efficientnetv2} & 480 & 55M & 24.0G  & 280.7 &  85.1 \\
& \textcolor{blueish}{$\bullet$}ConvNeXt-S~\cite{liu2022convnet} & 224 & 50M & 8.7G  & 447.1 & 83.1 \\
& \textcolor{blueish}{$\bullet$}ConvNeXt-B~\cite{liu2022convnet} & 224 & 89M & 15.4G  & 292.1 & 83.8 \\
& \textcolor{blueish}{$\bullet$}ConvNeXt-L~\cite{liu2022convnet} & 224 & 198M & 34.4G  & 146.8 & 84.3 \\
\midrule
\multirow{13}{*}{ViTs} & \textcolor{brickred}{$\circ$}ViT-B/32~\cite{dosovitskiy2020image} & 384 & 86M & 55.4G  &  85.9 & 77.9 \\
& \textcolor{brickred}{$\circ$}ViT-B/16~\cite{dosovitskiy2020image} & 384 & 307M & 190.7G & 27.3  & 76.5 \\
& \textcolor{brickred}{$\circ$}DeiT-B~\cite{touvron2021training} & 384 & 86M & 55.4G & 85.9 & 83.1 \\
& \textcolor{brickred}{$\circ$}CaiT-M24~\cite{touvron2021going} & 224 & 186M & 36.0G  & - & 83.4 \\
& \textcolor{brickred}{$\circ$}CaiT-M24~\cite{touvron2021going} & 384 & 186M & 116.1G & - & 84.5 \\
& \textcolor{brickred}{$\circ$}DeepViT-L~\cite{zhou2021deepvit} & 224 & 55M & 12.5G & - & 83.1 \\
& \textcolor{brickred}{$\circ$}T2T-ViT-24~\cite{yuan2021tokens} & 224 & 64M & 15.0G  & - & 82.6 \\
& \textcolor{brickred}{$\circ$}Swin-S~\cite{liu2021swin} & 224 & 50M & 8.7G & 436.9 &  83.0 \\
& \textcolor{brickred}{$\circ$}Swin-B~\cite{liu2021swin} & 384 & 88M & 47.0G & 84.7 &  84.5 \\
& \textcolor{brickred}{$\circ$}CSwin-B~\cite{dong2021cswin} & 224 & 78M & 15.0G & 250 & 84.2 \\
& \textcolor{brickred}{$\circ$}CSwin-B~\cite{dong2021cswin} & 384 & 78M & 47.0G & - & 85.4 \\
& \textcolor{brickred}{$\circ$}Focal-S~\cite{yang2021focal} & 224 & 51M & 9.1G & - & 83.5 \\
& \textcolor{brickred}{$\circ$}Focal-B~\cite{yang2021focal} & 224 & 90M & 16.0G & - & 83.8 \\
\midrule
\multirow{17}{*}{Hybrid} &
\textcolor{darkgreen}{$\diamond$}CvT-21~\cite{wu2021cvt} & 384 & 32M & 24.9G  & - & 83.3 \\
& \textcolor{darkgreen}{$\diamond$}CoAtNet-2~\cite{dai2021coatnet} & 224 & 75M & 15.7G  & 247.7 &  84.1 \\
& \textcolor{darkgreen}{$\diamond$}CoAtNet-3~\cite{dai2021coatnet} & 224 & 168M & 34.7G  & 163.3 & 84.5 \\
& \textcolor{darkgreen}{$\diamond$}CoAtNet-3~\cite{dai2021coatnet} & 384 & 168M & 107.4G &  48.5 & 85.8 \\
& \textcolor{darkgreen}{$\diamond$}CoAtNet-3~\cite{dai2021coatnet} & 512 & 168M & 203.1G &  22.4 & 86.0 \\
\cmidrule(lr){2-7}
& \textcolor{darkgreen}{$\diamond$}MaxViT-T  & 224 & 31M & 5.6G & 349.6 & 83.62 \\
& \textcolor{darkgreen}{$\diamond$}MaxViT-S  & 224 & 69M & 11.7G & 242.5 & 84.45 \\
& \textcolor{darkgreen}{$\diamond$}MaxViT-B & 224 & 120M & 23.4G & 133.6 & 84.95 \\
& \textcolor{darkgreen}{$\diamond$}MaxViT-L  & 224 &  212M & 43.9G & 99.4 & 85.17 \\
\cmidrule(lr){2-7}
& \textcolor{darkgreen}{$\diamond$}MaxViT-T  & 384 & 31M & 17.7G & 121.9 & 85.24 \\
& \textcolor{darkgreen}{$\diamond$}MaxViT-S & 384 & 69M & 36.1G & 82.7 & 85.74 \\
& \textcolor{darkgreen}{$\diamond$}MaxViT-B & 384 & 120M & 74.2G & 45.8 & 86.34\\
& \textcolor{darkgreen}{$\diamond$}MaxViT-L & 384 &  212M & 133.1G & 34.3 & 86.40 \\
\cmidrule(lr){2-7}
& \textcolor{darkgreen}{$\diamond$}MaxViT-T & 512 & 31M & 33.7G &  63.8 & 85.72 \\
& \textcolor{darkgreen}{$\diamond$}MaxViT-S & 512 & 69M & 67.6G & 43.3 & 86.19 \\
& \textcolor{darkgreen}{$\diamond$}MaxViT-B & 512 & 120M & 138.5G & 24.0 & 86.66 \\
& \textcolor{darkgreen}{$\diamond$}MaxViT-L & 512 &  212M & 245.4G & 17.8 & \textbf{86.70} \\
% \bottomrule
\end{tabular}
\end{table*}

\begin{table*}[t]
\footnotesize
\centering
\setlength{\tabcolsep}{4pt}
\renewcommand{\arraystretch}{0.9}
\caption{\normalsize \textbf{Performance comparison for large-scale data regimes}: ImageNet-21K and JFT pretrained models.}
\label{tab:imagenet21k-jft-comparison}
\begin{tabular}{c|l|cccccc}
% \hline
% \rowcolor[gray]{0.95}
& \multirow{2}{*}{Model} & \multirow{2}{*}{\begin{tabular}{@{}c@{}}Eval\\size\end{tabular}} & \multirow{2}{*}{Params} & \multirow{2}{*}{FLOPs} &
\multicolumn{2}{c}{IN-1K top-1 acc.} \\\cline{6-7}
% \vspace{0.5mm}
\rule{0pt}{2ex}& & & & & 21K$\rightarrow$1K & JFT$\rightarrow$1K\\ 
% \rowcolor[gray]{0.95}
\toprule
\multirow{7}{*}{ConvNets} &
\textcolor{blueish}{$\bullet$}BiT-R-101x3~\cite{kolesnikov2020big} & 384 & 388M & 204.6G & 84.4 & - \\
& \textcolor{blueish}{$\bullet$}BiT-R-152x4~\cite{kolesnikov2020big} & 480 & 937M & 840.5G & 85.4 & -\\
& \textcolor{blueish}{$\bullet$}EffNetV2-L~\cite{tan2021efficientnetv2} & 480 & 121M & 53.0G  &  86.8 & - \\
& \textcolor{blueish}{$\bullet$}EffNetV2-XL~\cite{tan2021efficientnetv2} & 512 & 208M & 94.0G &  87.3 & - \\
& \textcolor{blueish}{$\bullet$}ConvNeXt-L~\cite{liu2022convnet} & 384 & 198M & 101.0G &  87.5 & - \\
& \textcolor{blueish}{$\bullet$}ConvNeXt-XL~\cite{liu2022convnet} & 384 & 350M & 179.0G & 87.8 & -  \\
& \textcolor{blueish}{$\bullet$}NFNet-F4+~\cite{brock2021high} & 512 & 527M & 367G & - & 89.20  \\
\midrule
\multirow{7}{*}{ViTs} &
\textcolor{brickred}{$\circ$}ViT-B/16~\cite{dosovitskiy2020image} & 384 & 87M & 55.5G & 84.0 & - \\
& \textcolor{brickred}{$\circ$}ViT-L/16~\cite{dosovitskiy2020image} & 384 & 305M & 191.1G &  85.2 \\
& \textcolor{brickred}{$\circ$}ViT-L/16~\cite{dosovitskiy2020image} & 512 & 305M & 364G &  - & 87.76 \\
& \textcolor{brickred}{$\circ$}ViT-H/14~\cite{dosovitskiy2020image} & 518 & 632M & 1021G &  - & 88.55 \\
& \textcolor{brickred}{$\circ$}HaloNet-H4~\cite{vaswani2021scaling} & 512 & 85M & - &  85.8 & - \\
& \textcolor{brickred}{$\circ$}SwinV2-B~\cite{liu2021swin} & 384 & 88M & - &  87.1 & - \\
& \textcolor{brickred}{$\circ$}SwinV2-L~\cite{liu2021swin} & 384 & 197M & -& 87.7 & - \\
\midrule
\multirow{11}{*}{Hybrid} &
\textcolor{darkgreen}{$\diamond$}CvT-W24~\cite{wu2021cvt} & 384 & 277M & 193.2G &  87.7 & - \\
& \textcolor{darkgreen}{$\diamond$}R+ViT-L/16~\cite{dosovitskiy2020image} & 384 & 330M & - & - & 87.12 \\
& \textcolor{darkgreen}{$\diamond$}CoAtNet-3~\cite{dai2021coatnet} & 384 & 168M & 107.4G &  87.6 & 88.52 \\
& \textcolor{darkgreen}{$\diamond$}CoAtNet-3~\cite{dai2021coatnet} & 512 & 168M & 214G &  87.9 & 88.81 \\
& \textcolor{darkgreen}{$\diamond$}CoAtNet-4~\cite{dai2021coatnet} & 512 & 275M & 360.9G &  88.1 & 89.11 \\
& \textcolor{darkgreen}{$\diamond$}CoAtNet-5~\cite{dai2021coatnet} & 512 & 688M & 812G &  - & \textbf{89.77} \\
\cmidrule(lr){2-7}
& \textcolor{darkgreen}{$\diamond$}MaxViT-B & 384 & 119M & 74.2G  & 88.24 & 88.69 \\
& \textcolor{darkgreen}{$\diamond$}MaxViT-L & 384 & 212M & 128.7G & 88.32 & 89.12 \\
& \textcolor{darkgreen}{$\diamond$}MaxViT-XL & 384 & 475M & 293.7G & \textbf{88.51} &  89.36 \\
\cmidrule(lr){2-7}
& \textcolor{darkgreen}{$\diamond$}MaxViT-B & 512 & 119M & 138.3G & 88.38 &  88.82 \\
& \textcolor{darkgreen}{$\diamond$}MaxViT-L & 512 &  212M & 245.2G & 88.46 &  89.41 \\
& \textcolor{darkgreen}{$\diamond$}MaxViT-XL & 512 & 475M & 535.2G & \textbf{88.70} & \textbf{89.53}  \\
% \bottomrule
\end{tabular}
\end{table*}%

\subsection{Image Classification on ImageNet-1K}

\begin{figure}[!t]
 \centering
 \begin{subfigure}[b]{0.49\textwidth}
     \centering
     \includegraphics[width=\textwidth]{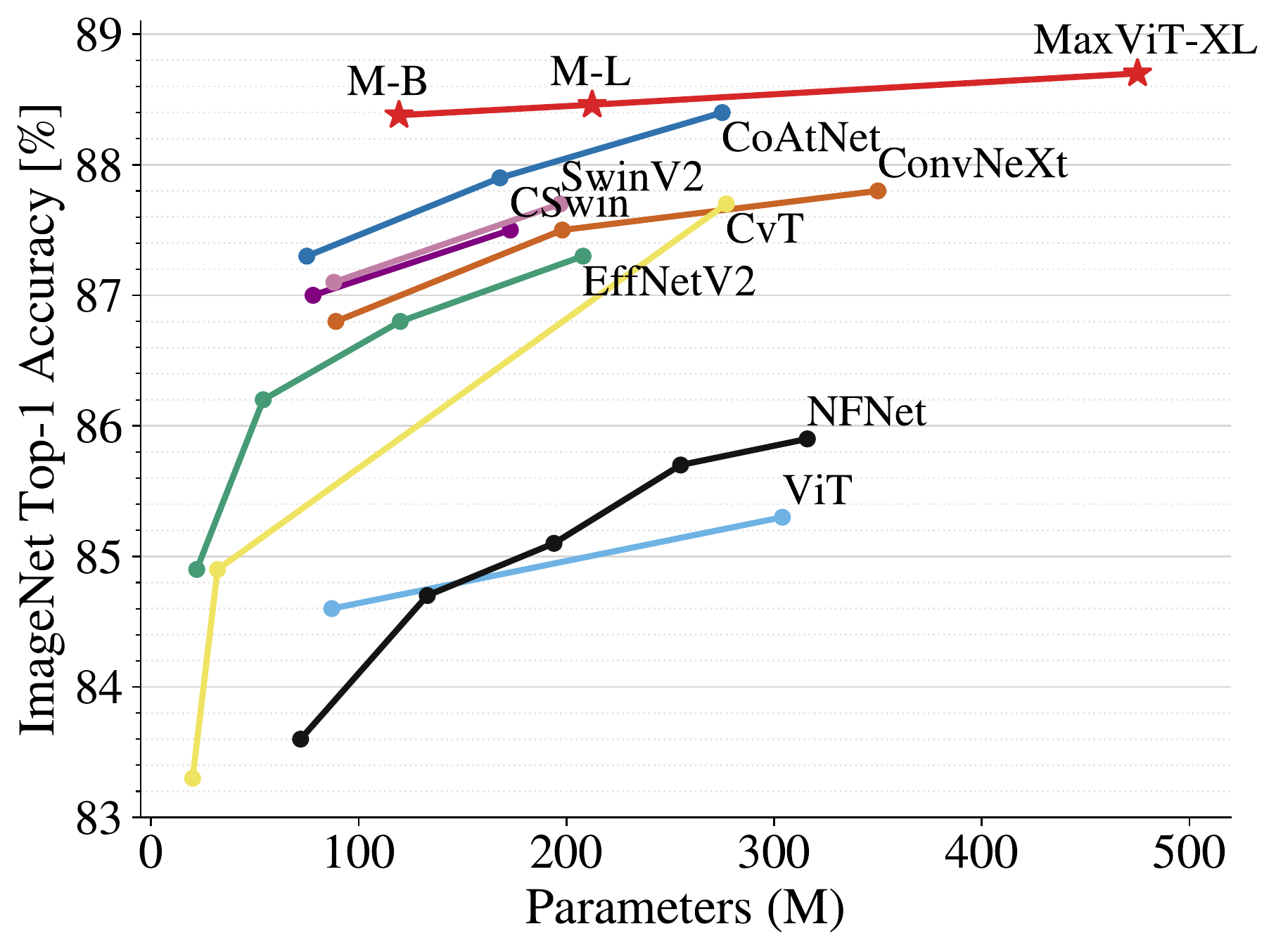}
     \caption{Accuracy \vs Params performances for ImageNet-21K pre-trained models.}
     \label{fig:imagenet-21k}
 \end{subfigure}
 \hfill
 \begin{subfigure}[b]{0.49\textwidth}
     \centering
     \includegraphics[width=\textwidth]{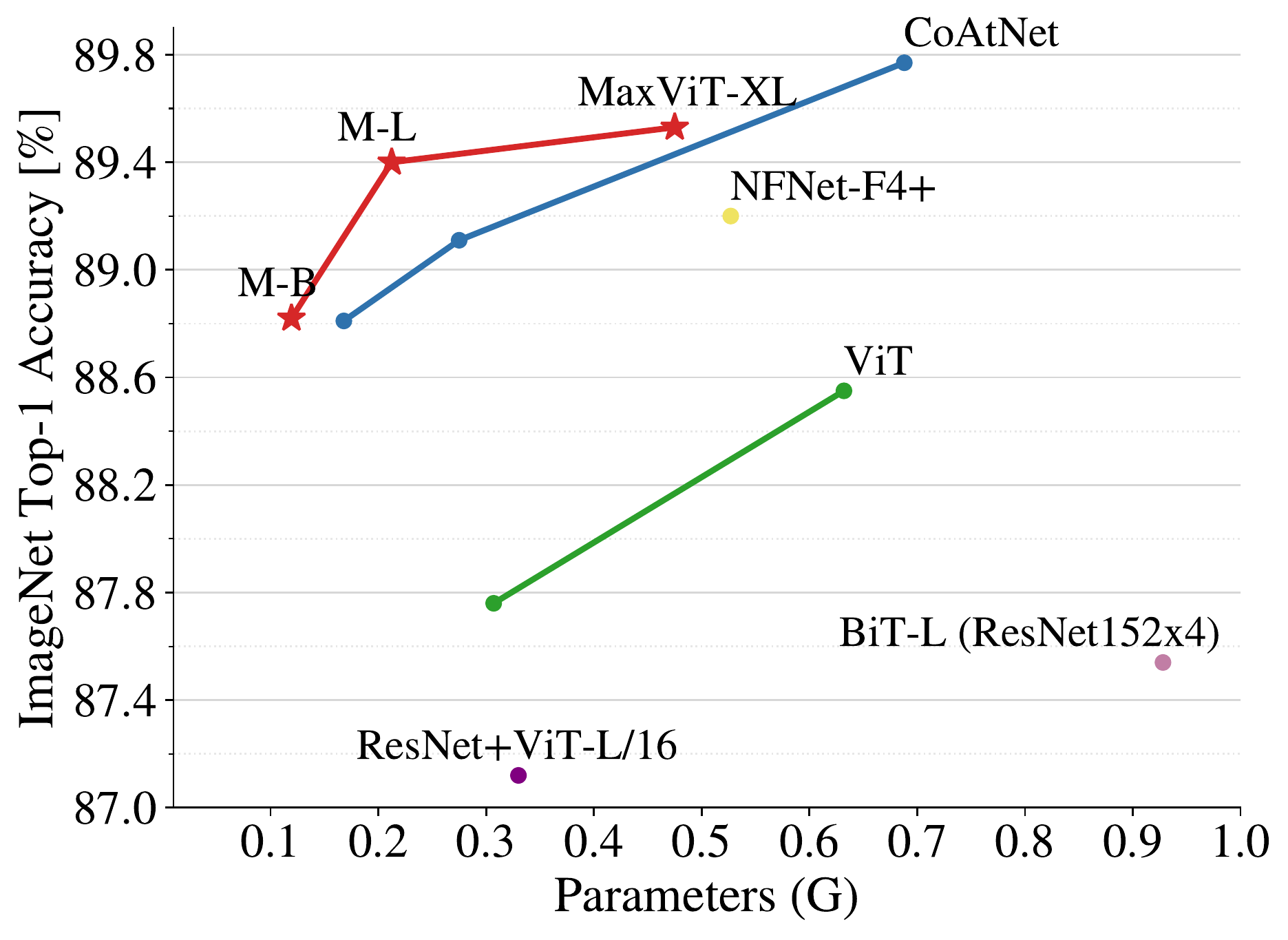}
     \caption{Accuracy \vs Params scaling curve for JFT-300M pre-trained models.}
     \label{fig:jft}
 \end{subfigure}
\caption{\textbf{Performance comparison on large-scale pre-trained models.} MaxViT shows superior scaling performance under both ImageNet-21K and JFT-300M pre-trained settings.}
\label{fig:large-scale-data-pt}
\end{figure}

\noindent\textbf{ImageNet-1K.} We show in Table~\ref{tab:imagenet1k-comparison} the performance comparisons on ImageNet-1K classification. Under the basic 224$\times$224 setting, MaxViT outperformed the most recent strong hybrid model CoAtNet by a large margin across the entire FLOPs spectrum, as shown in Figure~\ref{fig:imagenet-flops}. The MaxViT-L model sets a new performance record of 85.17\% at $224\times224$ training without extra training strategies, outperforming CoAtNet-3 by 0.67\%. In regards to throughput-accuracy trade-offs at $224^2$, MaxViT-S obtains 84.45\% top-1 accuracy, 0.25\% higher than CSWin-B and 0.35\% higher than CoAtNet-2 with comparable throughput.

When fine-tuned at higher resolutions (384/512), MaxViT continues to deliver high performance compared to strong ConvNet and Transformer competitors: (1) at $384^2$, MaxViT-B attains 86.34\% top-1 accuracy, outperforming EfficientNetV2-L by 0.64\%; (2) when fine-tuned at $512^2$, our MaxViT-L (212M) achieves top-1 accuracy 86.7\% , setting new SOTA performance on ImageNet-1K under the normal training setting. As Figure~\ref{fig:imagenet-1k-main} shows, MaxViT scales much better than SOTA vision Transformers on the ImageNet-1K trained model scale.

\hspace{-5.5mm}\textbf{ImageNet-21K.} Table~\ref{tab:imagenet21k-jft-comparison} shows the results of models pre-trained on ImageNet-21K. Remarkably, the MaxViT-B model achieves 88.38\% accuracy, outperforming the previous best model CoAtNet-4 by 0.28\% using only 43\% of parameter count and 38\% of FLOPs, demonstrating greater parameter and computing efficiency. Figure~\ref{fig:imagenet-21k} visualizes the model size comparison --  MaxViT scales significantly better than previous attention-based models of similar complexities, across the board. Additionally, the MaxViT-XL model achieves new SOTA performance, an accuracy of 88.70\% when fine-tuned at resolution $512\times512$.

\hspace{-5.5mm}\textbf{JFT-300M.} We also trained our model on a larger-scale proprietary dataset JFT-300M which contains $\sim$300 million weakly labeled images. As shown in Table~\ref{tab:imagenet21k-jft-comparison} and Figure~\ref{fig:jft}, our model is also scalable to massive scale training data -- MaxViT-XL achieves a high accuracy of 89.53\% with 475 million parameters, outperforming previous models under comparable model sizes. Due to resource limitations, we leave experiments on billion-parameter-scale models on planet-scale datasets (\eg JFT-3B~\cite{zhai2021scaling}) as future work.

% \hspace{-5.5mm}\textbf{Remarks on model throughput.} It is known that window partition operations are often implemented as a group of tensor reshapes are generally slow on dedicated hardware such as TPUs. Our MaxViT model involves the same number of window partition and reverse operations as Swin~\cite{liu2021swin} in each block, and is thus prone to potentially slower inference time. Enjoying greater model efficiency, however, MaxViT-S achieves 84.45\% accuracy, surpassing the previous best model CoAtNet-2 by 0.35\% while having similar throughput. We expect future processor improvements on more efficient local attention or reshaping operations to further boost the inference speed of MaxViT.

\subsection{Object Detection and Instance Segmentation}

\textbf{Setting.} 
We evaluated the MaxViT architectures on the COCO2017~\cite{lin2014microsoft} object bounding box detection and instance segmentation tasks with a two-stage framework~\cite{ren2015fasterrcnn}.
On the object detection task, a feature-pyramid architecture~\cite{Lin2017FeaturePN} was employed to boost different levels of objectiveness.
In the instance segmentation task, a well-known Cascade Mask-RCNN framework~\cite{he2017maskrcnn} was employed.
The dataset contains 118K training and 5K validation samples.
For all the compared models, the backbones are first pretrained using ImageNet-1K.
The pretrained models are then used to finetune on the detection and segmentation tasks.

\noindent\textbf{Results on COCO.} 
As shown in Table~\ref{tab:coco}, $AP$, $AP_{50}$, and $AP_{75}$ are reported for comparison.
The parameters and FLOPs are also reported as a reference for model complexity. 
The MaxViT backbone models, used in object detection and segmentation tasks, outperform all other backbones by large margins, including Swin, ConvNeXt, and UViT at various model sizes with respect to both accuracy and efficiency.
Note that MaxViT-S outperforms other base-level models (\eg Swin-B, UViT-B), with about 40\% less computational cost.

\begin{table}[!ht]
\centering
\footnotesize
\setlength{\tabcolsep}{2.5pt}
\renewcommand{\arraystretch}{1.}
\caption{\textbf{Comparison of two-stage object detection and instance segmentation on COCO2017.}
All models are pretrained on ImageNet-1K.}
\begin{tabular}{l|ccccccc|cc}
Backbone & Resolution & AP & AP$_{50}$ & AP$_{75}$ & AP$^{m}$ & AP$^{m}_{50}$ & AP$^{m}_{75}$ & FLOPs & Pars. \\
\toprule
\textcolor{blueish}{$\bullet$}ResNet-50~\cite{he2016deep} & 1280$\times$800 & 46.3 & 64.3 & 50.5 & 40.1 & 61.7 & 43.4 & 739G & 82M \\
\textcolor{blueish}{$\bullet$}X101-32~\cite{xie2017aggregated} & 1280$\times$800 & 48.1 & 66.5 & 52.4 & 41.6 & 63.9 & 45.2 & 819G & 101M \\
\textcolor{blueish}{$\bullet$}X101-64~\cite{xie2017aggregated} & 1280$\times$800 & 48.3 & 66.4 & 52.3 & 41.7 & 64.0 & 45.1 & 972G & 140M \\
\textcolor{blueish}{$\bullet$}ConvNeXt-T~\cite{liu2022convnet} & 1280$\times$800 & 50.4 & 69.1 & 54.8 & 43.7 & 66.5 & 47.3 &741G & -  \\
\textcolor{blueish}{$\bullet$}ConvNeXt-S~\cite{liu2022convnet} & 1280$\times$800 & 51.9 & 70.8 & 56.5 & 45.0 & 68.4 & 49.1 & 827G & - \\
\textcolor{blueish}{$\bullet$}ConvNeXt-B~\cite{liu2022convnet} & 1280$\times$800 & 52.7 & 71.3 & 57.2 & 45.6 & 68.9 & 49.5 & 964G & - \\
%\textcolor{blueish}{$\bullet$}ConvNeXt-L$^\ddagger$ & 1280$\times$800 & 54.8 & 73.8 & 59.8 & 47.6 & 71.3 & 51.7 & 1354 & -\\
\midrule
\textcolor{brickred}{$\circ$}Swin-T~\cite{liu2021swin} & 1280$\times$800 &  50.4 & 69.2 & 54.7 & 43.7 & 66.6 & 47.3 & 745G &  86M \\
\textcolor{brickred}{$\circ$}Swin-S~\cite{liu2021swin} & 1280$\times$800 & 51.9 & 70.7 & 56.3 & 45.0 & 68.2 & 48.8 & 838G & 107M \\
\textcolor{brickred}{$\circ$}Swin-B~\cite{liu2021swin} & 1280$\times$800 & 51.9 & 70.5 & 56.4 & 45.0 & 68.1 & 48.9 & 982G & 145M \\
%\textcolor{brickred}{$\circ$}Swin-L$^\ddagger$ & 1280$\times$800 & 53.9 & 72.4 & 58.8 & 46.7 & 70.1 & 50.8 & 1382 & - \\
%\textcolor{brickred}{$\circ$}Swin-L$^\ddagger$ & 1280$\times$800 & 53.9 & 72.4 & 58.8 & 46.7 & 70.1 & 50.8 & 1382 & - \\
\textcolor{brickred}{$\circ$}UViT-T~\cite{chen2022uvit} & 896$\times$896 & 51.1 & 70.4 & 56.2 & 43.6 & 67.7 & 47.2 & 613G & 47M \\ % https://tensorboard.corp.google.com/experiment/2533535615840984763/#timeseries
\textcolor{brickred}{$\circ$}UViT-S~\cite{chen2022uvit} & 896$\times$896 & 51.4 & 70.8 & 56.2 & 44.1 & 68.2 & 48.0 & 744G & 54M \\ %https://tensorboard.corp.google.com/experiment/2187330177719871319/#timeseries
\textcolor{brickred}{$\circ$}UViT-B~\cite{chen2022uvit} & 896$\times$896 & 52.5 & 72.0 & 57.6 & 44.3 & 68.7 & 48.3 & 975G & 74M \\ % https://tensorboard.corp.google.com/experiment/4467822036480772036/#timeseries
\textcolor{brickred}{$\circ$}As-ViT-L~\cite{chen2022autoscaling} & 1024$\times$1024 & 52.7 & 72.3 & 57.9 & 45.2 & 69.7 & 49.8 & 1094G & 139M \\ %https://tensorboard.corp.google.com/experiment/7624284926121512164/#timeseries

\midrule
\textcolor{darkgreen}{$\diamond$}MaxViT-T & 896$\times$896 & 52.1 & 71.9 & 56.8 & 44.6 & 69.1 & 48.4 & 475G & 69M\\
\textcolor{darkgreen}{$\diamond$}MaxViT-S & 896$\times$896 & 53.1 & 72.5 & 58.1 & 45.4 & 69.8 & 49.5 & 595G & 107M\\
\textcolor{darkgreen}{$\diamond$}MaxViT-B & 896$\times$896 & \textbf{53.4} & \textbf{72.9} & \textbf{58.1} & \textbf{45.7} & \textbf{70.3} & \textbf{50.0} & 856G & 157M \\
% \bottomrule
\end{tabular}
\label{tab:coco}
\end{table}

\subsection{Image Aesthetic Assessment.}

\textbf{Setting.} We train and evaluate the MaxViT model on the AVA benchmark~\cite{murray2012ava} which contains 255K images with aesthetics scores rated by amateur photographers. Similar to \cite{talebi2018nima}, we split the dataset into 80\%/20\% training and test sets. We followed~\cite{talebi2018nima} and used the normalized Earth Mover's Distance as our training loss. We trained MaxViT at three different input resolutions: $224^2$, $384^2$ and $512^2$, initialized with ImageNet-1K pre-trained weights.

\noindent\textbf{Results on AVA.} To evaluate and compare our model against existing methods, we present a summary of our results in Table~\ref{tab:iqa-comparison}. For similar input resolutions, the proposed MaxViT-T model outperforms existing image aesthetic assessment methods. As the input resolution increases, the performance improves, benefiting from its strong non-local capacity. Also, MaxViT shows better linear correlation compared to the SOTA method~\cite{ke2021musiq} which uses multi-resolution inputs.

\subsection{Image Generation}
\label{ssec:gan}

\textbf{Setting.} We evaluate the generative ability of MaxViT blocks to generate images of 128x128 resolution on ImageNet-1K. We choose the unconditional image generation to focus on the performance of different generators in GANs. We use the Inception Score (IS)~\cite{salimans2016improved} and the Fr{\'e}chet Inception Distance (FID)~\cite{heusel2017gans} as quantitative evaluation metrics. 50,000 samples were randomly generated to calculate the FID and IS scores.
We compared MaxViT against HiT~\cite{zhao2021improved}, a SOTA generative Transformer model, which uses attention at low resolutions (e.g., 32, 64), and using implicit neural functions at high resolutions (e.g., 128). By contrast, MaxViT uses the proposed MaxViT block at every resolution. Note that we use an inverse block order (\texttt{GA-BA-Conv}) as we found it to perform better (see Table~\ref{tab:ablation-order}). Since Batch Normalization~\cite{ioffe2015batch,zhao2021improved} achieves better results on image generation, we replaced all Layer Norm with Batch Norm under this setting.

\noindent\textbf{Results on ImageNet-1K.} The results are shown in Table~\ref{tab:generation-comparison}. Our MaxViT achieved better FID and IS with significantly lower number of parameters. These results demonstrate the effectiveness of MaxViT blocks for generation tasks. More details of the generative experiment can be found in Appendix.

\begin{table*}[!t]
\centering
\setlength{\tabcolsep}{3pt}
\begin{tabular}{@{}cc@{}}
\begin{minipage}[t]{0.57\textwidth}
% \centeringbat
\setlength{\tabcolsep}{1.6pt}
\renewcommand{\arraystretch}{1.0}
\caption{\textbf{Image aesthetic assessment results on the AVA benchmark~\cite{murray2012ava}}. PLCC and SRCC represent the Pearson's linear and Spearman's rank correlation coefficients.}
\label{tab:iqa-comparison}
\begin{tabular}{l|cccc}
% \hline
% \rowcolor[gray]{0.95}
Model & Res. & Pars. & PLCC$\uparrow$& SRCC$\uparrow$ \\
\toprule
\textcolor{blueish}{$\bullet$}NIMA~\cite{talebi2018nima} & $224$ & 56M & 0.636 & 0.612 \\
\textcolor{blueish}{$\bullet$}EffNet-B0~\cite{tan2019efficientnet} & $224$ & 5.3M & 0.642 & 0.620 \\
\textcolor{blueish}{$\bullet$}AFDC\cite{chen2020adaptive} & $224$ & 44.5M & 0.671 & 0.649 \\\hline
\textcolor{brickred}{$\circ$}ViT-S/32~\cite{ke2021musiq} & $384$ & 22M & 0.665 & 0.656 \\
\textcolor{brickred}{$\circ$}ViT-B/32~\cite{ke2021musiq} & $384$ & 88M & 0.664 & 0.664 \\
\textcolor{brickred}{$\circ$}MUSIQ~\cite{ke2021musiq} & \footnotesize{$224\!\sim\!512$} & 27M & 0.720 & 0.706 \\\hline
%\textcolor{brickred}{$\circ$}MUSIQ~\cite{ke2021musiq} & Full-res & 27M & 0.731 & \textbf{0.719} \\\hline
\textcolor{darkgreen}{$\diamond$}MaxViT-T & $224$ & 31M & 0.707 & 0.685 \\
\textcolor{darkgreen}{$\diamond$}MaxViT-T & $384$ & 31M & 0.736 & 0.699 \\
\textcolor{darkgreen}{$\diamond$}MaxViT-T & $512$ & 31M & \textbf{0.745} & \textbf{0.708} \\
% \bottomrule
\end{tabular}
\end{minipage}%
&
% \hspace{0.5mm}
\begin{minipage}[t]{0.4\textwidth}
% \centeringbat
\setlength{\tabcolsep}{2.5pt}
\renewcommand{\arraystretch}{1.}
\caption{\textbf{Comparison of image generation on ImageNet}. $\ddagger$ used a pre-trained ImageNet classifier.}
\label{tab:generation-comparison}
\begin{tabular}{l|cc}
% \hline
% \rowcolor[gray]{0.95}
Model  & FID$\downarrow$& IS$\uparrow$ \\
\toprule
\textcolor{blueish}{$\bullet$}GAN~\cite{goodfellow2014generative}  & 54.17 & 14.01 \\
\textcolor{blueish}{$\bullet$}PacGAN2~\cite{lin2018pacgan}  & 57.51 & 13.50 \\
\textcolor{blueish}{$\bullet$}MGAN~\cite{hoang2018mgan}  & 50.90 & 14.44 \\
\textcolor{blueish}{$\bullet$}LogoGAN~\cite{sage2018logo}$\ddagger$  & 38.41 & 18.86 \\
% Logo-GAN-RC~\cite{} \\
\textcolor{blueish}{$\bullet$}SS-GAN~\cite{chen2019self}  & 43.87 & - \\
\textcolor{blueish}{$\bullet$}SC GAN~\cite{liu2020diverse}  & 40.30 & 15.82  \\
\textcolor{blueish}{$\bullet$}ConvNet-$R_1$~\cite{zhao2021improved} & 37.18 & 19.55 \\ \hline
\textcolor{brickred}{$\circ$}HiT~\cite{zhao2021improved}~(32.9M) & 30.83 & 21.64 \\\hline
\textcolor{darkgreen}{$\diamond$}MaxViT~(18.6M) & \textbf{30.77} & \textbf{22.58} \\
% \bottomrule
\end{tabular}
\end{minipage}%
\end{tabular}
\end{table*}

% \begin{table*}[!t]
% \centering
% \setlength{\tabcolsep}{1pt}
% \begin{tabular}{@{}cc@{}}
% \begin{minipage}[t]{0.48\textwidth}
% % \centeringbat
% \setlength{\tabcolsep}{3pt}
% \renewcommand{\arraystretch}{1.}
% \begin{tabular}{l|ccc}
% % \hline
% % \rowcolor[gray]{0.95}
% Model & Pars. & FID $\downarrow$& IS $\uparrow$ \\
% \toprule

% HiT~\cite{zhao2021improved} & 32.9M & 30.83 & 21.64 \\\hline
% MaxViT & 18.6M & \textbf{30.77} & \textbf{22.58} \\
% % \bottomrule
% \end{tabular}
% \caption{Comparison for unconditional image generation on ImageNet-1K}
% \label{tab:generation-comparison}
% \end{minipage}%
% &
% \end{tabular}
% \end{table*}

\subsection{Ablation Studies.}

In this section, we ablate important design choices in MaxViT on ImageNet-1K image classification. We use the MaxViT-T model trained for 300 epochs by default and report top-1 accuracy on ImageNet-1K. Except for the ablated design choice, we used the same training configurations, unless stated otherwise.

%% Not doable!
% \noindent\textbf{Resolution adaptability.} \textcolor{red}{Refer to CycleMLP. Better send an email asking for the datapoints. This is important since we want to show the benefits of being fully-conv.}

\begin{table*}[!t]
\centering
\setlength{\tabcolsep}{4pt}
\begin{tabular}{@{}cc@{}}
\begin{minipage}[t]{0.48\textwidth}
\centering
\setlength{\tabcolsep}{3pt}
\renewcommand{\arraystretch}{1.08}
\caption{\textbf{Effects of global grid-attention.} Ablate-S1 means we remove grid-attention in stage 1 while Replace-S1 means replacing grid-attention with block-attention.}
\label{tab:ablation-grid-attention}
\begin{tabular}{l|ccc}
% \hline
% \rowcolor[gray]{0.95}
Model & Pars. & FLOPs & Top-1 Acc. \\
\toprule
MaxViT-T & 30.9M & 5.6G & 83.62 \\\hline
 Ablate-S1 & 30.8M & 5.3G & 83.36\scriptsize\color{gray}{(-0.26)} \\
 Ablate-S2 & 30.5M & 5.3G & 83.38\scriptsize\color{gray}{(-0.24)}\\
 Ablate-S3 & 26.9M & 4.9G & 83.00\scriptsize\color{gray}{(-0.62)}\\\hline
 Replace-S1 & 30.9M & 5.6G & 83.49\scriptsize\color{gray}{(-0.13)}\\
 Replace-S2 & 30.9M & 5.6G & 83.41\scriptsize\color{gray}{(-0.22)}\\
 Replace-S3 & 30.9M & 5.6G & 83.40\scriptsize\color{gray}{(-0.23)}\\
% \bottomrule
\end{tabular}
\end{minipage}%
&
% \hspace{0.3mm}

\begin{minipage}[t]{0.48\textwidth}
\centering
\setlength{\tabcolsep}{3pt}
\renewcommand{\arraystretch}{0.91}
\caption{\textbf{Block order study.} C, BA, GA represent MBConv, block-, and grid-attention respectively.}
\label{tab:ablation-order}
\begin{tabular}{l|ccc}
% \hline
% \rowcolor[gray]{0.95}
Model & Pars. & FLOPs & Top-1 acc. \\
\toprule
C-BA-GA & 30.9M & 5.6G & 83.62 \\\hline
C-GA-BA & 30.9M & 5.6G & 83.54\scriptsize\color{gray}{(-0.08)} \\
BA-C-GA & 31.1M & 5.3G & 83.07\scriptsize\color{gray}{(-0.55)} \\
BA-GA-C & 31.1M & 5.3G & 83.02\scriptsize\color{gray}{(-0.60)} \\
GA-C-BA & 31.1M & 5.3G & 83.08\scriptsize\color{gray}{(-0.54)} \\
GA-BA-C & 31.1M & 5.3G & 83.03\scriptsize\color{gray}{(-0.59)} \\
\midrule
\multicolumn{4}{c}{GAN experiments} \\
Model & Pars. & FID$\downarrow$ & IS$\uparrow$ \\ \toprule
GA-BA-C & 18.6M & 30.77 & 22.68 \\ \hline
C-BA-GA & 18.6M & 31.40 & 21.49\scriptsize\color{gray}{(-1.19)}
% \bottomrule
\end{tabular}
\end{minipage}
\\
\begin{minipage}[t]{0.48\textwidth}
\centering
\setlength{\tabcolsep}{3pt}
\renewcommand{\arraystretch}{1.}
\caption{\textbf{Ablation of MBConv.} Ablate-S1 means we delete MBConv layers in stage 1. Note that the network will also be smaller if we ablate MBConv layers in some stage.}
\label{tab:ablation-mbconv}
\begin{tabular}{l|ccc}
% \hline
% \rowcolor[gray]{0.95}
Model & Pars. & FLOPs & Top-1 acc. \\
\toprule
MaxViT-T & 30.9M & 5.6G & 83.62 \\\hline
Ablate-S1 & 30.8M & 5.2G & 83.24\scriptsize\color{gray}{(-0.38)} \\
Ablate-S2 & 30.5M & 5.4G & 83.02\scriptsize\color{gray}{(-0.60)} \\
Ablate-S3 & 27.6M & 5.1G & 82.65\scriptsize\color{gray}{(-0.97)} \\
Ablate-S4 & 25.7M & 5.4G & 83.09\scriptsize\color{gray}{(-0.53)} \\
% \bottomrule
\end{tabular}
\end{minipage}%
&
% \hspace{0.3mm}

\begin{minipage}[t]{0.48\textwidth}
\centering
\setlength{\tabcolsep}{2.5pt}
\renewcommand{\arraystretch}{0.9}
\caption{\textbf{Sequential \vs parallel.} We compared our model with modified parallel multi-axis scheme $Paral$-$\star$.}
\label{tab:ablation-parallel}
\begin{tabular}{l|ccc}
% \hline
% \rowcolor[gray]{0.95}
Model & Pars. & FLOPs & Top-1 acc. \\
\toprule
MaxViT-T & 30.9M & 5.6G & 83.62 \\
$Paral$-T & 34.5M & 6.2G & 82.64\scriptsize\color{gray}{(-0.98)} \\ \hline
MaxViT-S & 68.9M & 11.7G & 84.45 \\ 
$Paral$-S & 76.9M & 13.0G & 83.45\scriptsize\color{gray}{(-1.00)} \\ \hline
MaxViT-B & 119.4M & 24.2G & 84.95  \\
$Paral$-B & 133.4M & 26.9G & 83.70\scriptsize\color{gray}{(-1.25)}\\ \hline
MaxViT-L & 211.8M & 43.9G & 85.17 \\
$Paral$-L & 236.6M & 48.8G & 83.54\scriptsize\color{gray}{(-1.63)} \\ 
% \bottomrule
\end{tabular}
\end{minipage}%
\\
\end{tabular}
\end{table*}

\noindent\textbf{Global grid-attention.} One of our main contributions is the grid-attention module, which allows for sparse global interactions at linear time, enabling our model to capture global information at all stages. We conducted two ablations to understand its gain: 1) completely removed global attention at each stage; 2) replaced grid attention with block attention to retain the same parameter count and FLOPs. As Table~\ref{tab:ablation-grid-attention} shows, enabling global attention at earlier stages can further boost performance over using only local attention or convolutions. 

\noindent\textbf{MBConv layer.} We also ablated the usage of MBConv layers in MaxViT by removing all MBConv in each stage. Note that we should also consider the reduction of parameter count and FLOPs when removing the MBConv layers. Plus, Stage 3 has 5 blocks whereas other stages have only 2. As Table~\ref{tab:ablation-mbconv} shows, the usage of MBConv layers in MaxViT significantly boosts performance.

\noindent\textbf{Block order study.} We present three different modules to build the MaxViT block -- MBConv, block-, and grid-attention -- which captures spatial interactions from local to global. To investigate the most effective way to combine them, we evaluated the MaxViT-T model using all 6 permutations. We always apply downsampling in the first layer, which might cause a minor model size difference. We can observe from Table~\ref{tab:ablation-order} that placing MBConv before attention layers is almost always better than other combinations.
The reason might be that it is more suitable to get local features/patterns in early layers, then aggregate them globally, which is aligned with existing hybrid models~\cite{dai2021coatnet,xiao2021early}, which puts Conv layers in front of attention.
In generative experiments (Section~\ref{ssec:gan}), however, we found the best order to be from global to local: \texttt{GA-BA-C}. We hypothesize that it may be advantageous for generation tasks to first obtain the overall structures correct with global processing blocks (\ie grid-attention layers), then fill in finer details using local processing blocks (\ie MBConv).

\begin{wrapfigure}{R}{0.4\textwidth}
\vspace{-11mm}
\begin{center}
\includegraphics[width=0.4\textwidth]{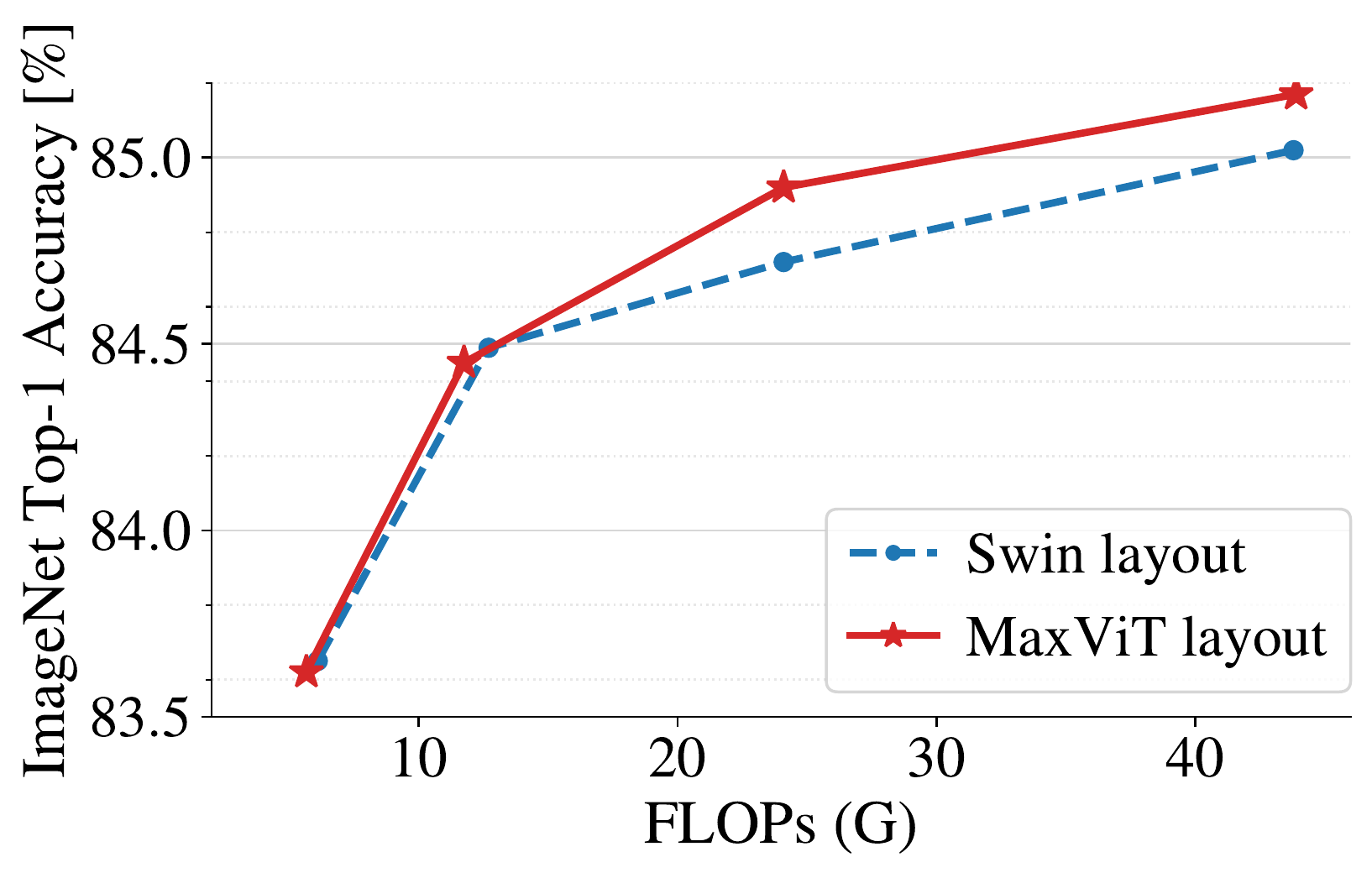}
\end{center}
\vspace{-6mm}
\caption{Vertical layout ablation. Our model scales better than Swin layeout~\cite{liu2021swin}.}
\label{fig:ablation-layout}
\vspace{-12mm}
\end{wrapfigure}

\noindent\textbf{Sequential \vs parallel.} In our approach, we sequentially stack the multi-axis attention modules following~\cite{liu2021swin,wang2020axial}, while there also exist other models that adopt a parallel design~\cite{zhao2021improved,tu2022maxim}. In this ablation, we compare our sequential Max-SA against parallel branches containing block- and grid-attention respectively. Note that we use an input projection to double the channels, then split the heads to feed the two branches in order to remain similar complexity to MaxViT, and an output projection that reduces the concatenated branches. We did rough parameter tuning and found that an initial learning rate of $10^{-3}$ performs significantly better than $3\times10^{-3}$ for parallel models. We use all the same parameters except the learning rate. As Table~\ref{tab:ablation-parallel} shows, our sequential approach remarkably outperforms parallel counterparts with fewer parameters and computation. The reason may be that the parallel designs learn complementary cues with less interactions between them, whereas our sequential stack is able to learn more powerful fusions between local and global layers.

\noindent\textbf{Vertical layout.} We further examine our vertical layout design, \ie the number of blocks each stage. We compared our design against the choice of Swin/ConvNeXt~\cite{liu2021swin,liu2022convnet}. We change MaxViT-T and -S to blocks $B=(2,2,6,2)$, and MaxViT-B, -L to have blocks $B=(2,2,18,2)$ strictly following the stage ratio of Swin~\cite{liu2021swin}. It may be seen from Figure~\ref{fig:ablation-layout} that our layout performed comparably to Swin for small models, but scales significantly better for larger models.

% \subsection{Attention Visualization. (Optional)}

\section{Discussion and Conclusion}
While recent works in the 2020s have arguably shown that ConvNets and vision Transformers can achieve similar performance on image recognition, our work presents a unified design that takes advantages of the best of both worlds -- efficient convolution and sparse attention -- and demonstrates that a model built on top, namely MaxViT, can achieve state-of-the-art performance on a variety of vision tasks, and more importantly, scale extremely well to massive scale data sizes.
%
% We propose a new vision Transformer called MaxViT based on an efficient multi-axis attention that can capture both local and global interactions in linear time.
%
% Equipped with a hybrid design of fused convolution and attention, MaxViT achieves state-of-the-art performance on a variety of vision tasks under a spectrum of data sizes, demonstrating its significant potential to be employed as a universal vision backbone.
%
Even though we present our model in the context of vision tasks, the proposed multi-axis approach can easily extend to language modeling to capture both local and global dependencies in linear time.
We also look forward to studying other forms of sparse attention in higher-dimensional or multi-modal signals such as videos, point clouds, and vision-languages.

\noindent\textbf{Societal impact.}
Investigating the performance and scalability of large model designs would consume considerable computing resources. These efforts can contribute to increased carbon emissions, which could hence raise environmental concerns. However, the proposed model offers strong modular candidates that expand the network's design space for future efforts on automated architectural design. If trained improperly, the proposed model may express bias and fairness issues. The proposed generative model can be abused to generate misleading media and fake news. These issues demand caution in future related research.

\noindent\textbf{Acknowledgment.}
We thank Xianzhi Du and Wuyang Chen for extensive help on experiments. We also thank Hanxiao Liu, Zihang Dai, Anurag Arnab, Huiwen Chang, Junjie Ke, Mauricio Delbracio, Sungjoon Choi, and Irene Zhu for valuable discussions and help.

%%%%%% For Arxiv!!!!
% \clearpage
\appendix
\section*{Appendix}

In this Appendix we provide the following material:

\begin{itemize}
\item Sec.~\ref{sec:model-details} describes the detailed architectures of MaxViT for image classification (Sec.~\ref{ssec:main-model}), object detection and segmentation (Sec.~\ref{ssec:detection-model}), image aesthetics assessment (Sec.~\ref{ssec:image-quality-model}), and image generation (Sec.~\ref{ssec:gan-model}). 
\item Sec.~\ref{sec:experimental-settings} presents complete training settings and hyperparameters for image classification (Sec.~\ref{ssec:imagenet-settings}), object detection and segmentation (Sec.~\ref{ssec:coco-settings}), image aesthetics assessment (Sec.~\ref{ssec:image-quality-settings}), and image generation (Sec.~\ref{ssec:gan-settings}). 
\item Sec.~\ref{sec:complete-experimental-results} demonstrates comprehensive experimental results, including image classification on ImageNet-1K (Table~\ref{tab:imagenet-1k-complete}), ImageNet-21K and JFT (Table~\ref{tab:imagenet21k-jft-complete}), as well as more image generation visualizations on ImageNet-1K (Figure~\ref{fig:more-gan-results}).

\end{itemize}

\section{Model Details}
\label{sec:model-details}

\subsection{Backbone Details}
\label{ssec:main-model}

\subsubsection{MBConv}
MaxViT leverages the MBConv block~\cite{sandler2018mobilenetv2,tan2019efficientnet} as the main convolution operator. We also adopt a pre-activation structure~\cite{he2016identity,dai2021coatnet} to promote homogeneity between MBConv and Transformer blocks. Specifically, assume $\mathbf{x}$ to be the input feature, the MBConv block without downsampling is formulated as:
\begin{equation}
\label{eq:mbconv}
\mathbf{x}\leftarrow \mathbf{x}+ \mathsf{Proj}(\mathsf{SE}(\mathsf{DWConv}(\mathsf{Conv}(\mathsf{Norm}(\mathbf{x}))))),
\end{equation}
\sloppy where $\mathsf{Norm}$ is $\mathsf{BatchNorm}$~\cite{ioffe2015batch}, $\mathsf{Conv}$ is the expansion Conv1x1 followed by $\mathsf{BatchNorm}$ and $\mathsf{GELU}$~\cite{hendrycks2016gaussian} activation, a typical choice for Transformer-based models. $\mathsf{DWConv}$ is the Depthwise Conv3x3 followed by $\mathsf{BatchNorm}$ and $\mathsf{GELU}$. $\mathsf{SE}$ is the Squeeze-Excitation layer~\cite{hu2018squeeze}, while $\mathsf{Proj}$ is the shrink Conv1x1 to down-project the number of channels. Note that for the first MBConv block in every stage, the downsampling is done by applying stride-2 Depthwise Conv3x3 while the shortcut branch should also apply pooling and channel projection:
\begin{equation}
\label{eq:mbconv-downsampling}
\mathbf{x}\leftarrow \mathsf{Proj}(\mathsf{Pool2D}(\mathbf{x}))+ \mathsf{Proj}(\mathsf{SE}(\mathsf{DWConv}\!\downarrow\!(\mathsf{Conv}(\mathsf{Norm}(\mathbf{x}))))).
\end{equation}

\subsubsection{Relative Attention}
Relative attention has been explored in several previous studies for both NLP~\cite{shaw2018self,wu2019pay} and vision~\cite{liu2021swin,vaswani2021scaling,dai2021coatnet,jiang2021transgan}.
Here to simplify the presentation, we present our model using only a single head of the multi-head self-attention.
In the actual implementation, we always use multi-head attention with the same head dimension.
The relative attention can be defined as:
\begin{equation}
\label{eq:relative-attention}
\mathsf{RelAttention}(Q,K,V)=\mathsf{softmax}(QK^T/\sqrt{d}+B)V,
\end{equation}
where $Q,K,V\in \mathbb{R}^{(H\times W)\times C}$ are the query, key, and value matrices and $d$ is the hidden dimension.
The attention weights are co-decided by a learned static location-aware matrix $B$ and the scaled input-adaptive attention $QK^T/\sqrt{d}$.
Considering the differences in 2D coordinates, the relative position bias $B$ is parameterized by a matrix $\hat{B}\in\mathbb{R}^{(2H-1)(2W-1)}$.
Following typical practices~\cite{liu2021swin,dai2021coatnet}, when fine-tuned at a higher resolution \eg $H'\times W'$, we use bilinear interpolation to map the relative positional bias from $\mathbb{R}^{(2H-1)(2W-1)}$ to $\mathbb{R}^{(2H'-1)(2W'-1)}$.
This relative attention benefits from input-adaptivity, translation equivariance, and global interactions, which is a preferred choice over the vanilla self-attention on 2D vision tasks.
In our model, all the attention operators use this relative attention defined in Eq.~\ref{eq:relative-attention} by default.

\subsubsection{Multi-Axis Attention}
We assume the relative attention operator in Eq.~\ref{eq:relative-attention} follows the convention for 1D input sequences \ie always regards the \textit{second last dimension} of an input $(..., L, C)$ as the \textit{spatial axis} where ${L},C$ represent sequence length and channels.
The proposed Multi-Axis Attention can be implemented without modification to the self-attention operation.
To start with, we first define the $\mathsf{Block}(\cdot)$ operator with parameter $P$ as partitioning the input image/feature $\mathbf{x}\in\mathbb{R}^{H\times W\times C}$ into non-overlapping blocks with each block having size $P\times P$.
Note that after window partition, the block dimensions are gathered onto the spatial dimension (\ie -2 axis):
\begin{equation}
\label{eq:block-operator}
\mathsf{Block}: (H,W,C)\rightarrow(\frac{H}{P}\times P,\frac{W}{P}\times P,C)\rightarrow (\frac{HW}{P^2},P^2,C).
\end{equation}
We denote the $\mathsf{Unblock}(\cdot)$ operation as the reverse of the above block partition procedure.
Similarly, we define the $\mathsf{Grid}(\cdot)$ operation with parameter $G$ as dividing the input feature into a uniform $G\times G$ grid, with each lattice having \textit{adaptive size} $\frac{H}{G}\times \frac{W}{G}$. Unlike the $\mathsf{block}$ operator, we need to apply an extra $\mathsf{Transpose}$ to place the grid dimension in the assumed spatial axis (\ie -2 axis):
\begin{equation}
\label{eq:grid-operator}
\mathsf{Grid}:(H,W,C)\rightarrow(G\times \frac{H}{G},G\times\frac{W}{G},C)\rightarrow \underbrace{(G^2,\frac{HW}{G^2},C)\rightarrow
(\frac{HW}{G^2},G^2,C)}_{\text{swapaxes(axis1=-2,axis2=-3)}}
\end{equation}
with its inverse operation $\mathsf{Ungrid}(\cdot)$ that reverses the gridded input back to the normal 2D feature space.

To this end, we are ready to explain the multi-axis attention module. Given an input tensor $\mathbf{x}\in\mathbb{R}^{H\times W\times C}$, the local Block Attention can be expressed as:
\begin{align}
\begin{split}
\label{eq:multi-axis-attention}
\mathbf{x}&\leftarrow \mathbf{x} + \mathsf{Unblock}(\mathsf{RelAttention}(\mathsf{Block}(\mathsf{LN}(\mathbf{x})))) \\ 
\mathbf{x}&\leftarrow\mathbf{x}+\mathsf{MLP}(\mathsf{LN}(\mathbf{x})) \\
\end{split}
\end{align}
while the global, dilated Grid Attention module is formulated as:
\begin{align}
\begin{split}
\mathbf{x}&\leftarrow \mathbf{x} + \mathsf{Ungrid}(\mathsf{RelAttention}(\mathsf{Grid}(\mathsf{LN}(\mathbf{x})))) \\ 
\mathbf{x}&\leftarrow\mathbf{x}+\mathsf{MLP}(\mathsf{LN}(\mathbf{x})) \\
\end{split}
\end{align}
where we omit the $QKV$ input format in the $\mathsf{RelAttention}$ operation for simplicity. $\mathsf{LN}$ denotes the Layer Normalization~\cite{ba2016layer}, where $\mathsf{MLP}$ is a standard MLP network~\cite{dosovitskiy2020image,liu2021swin} consisting of two linear layers: $\mathbf{x}\leftarrow W_2\mathsf{GELU}(W_1\mathbf{x})$.

\subsubsection{Comparison to Axial attention}
\label{sssec:comparison-axial}

\begin{wrapfigure}{r}{0.5\textwidth}
\begin{center}
% \vspace{-1mm}
\includegraphics[width=0.48\textwidth]{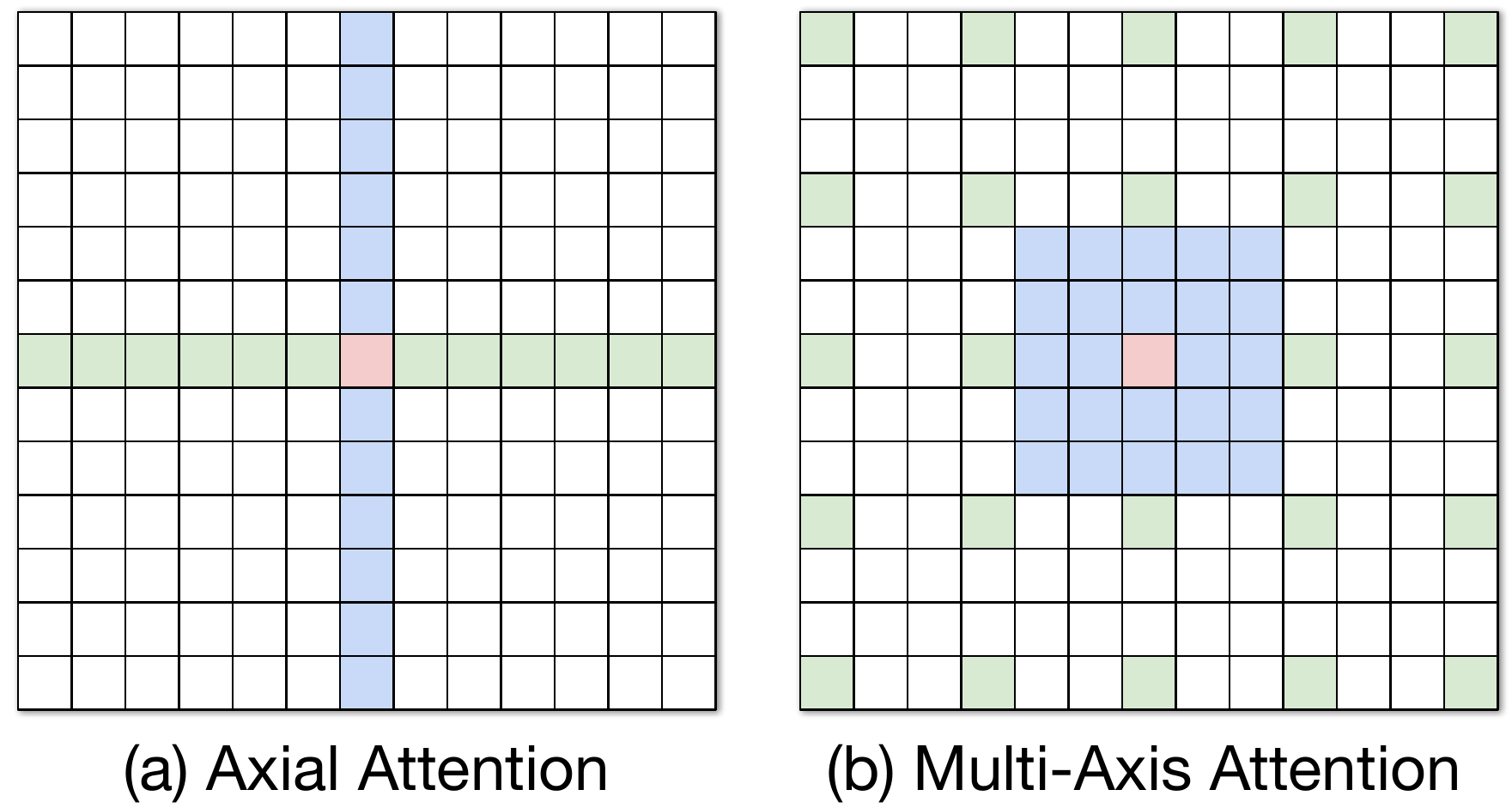}
\end{center}
\vspace{-2mm}
\caption{Comparison of Axial attention and our proposed Multi-Axis attention.}
\label{fig:axial-compare}
\vspace{-4mm}
\end{wrapfigure}

It should be noted that our proposed multi-axis attention (Max-SA) module is completely different from the axial attention proposed in~\cite{ho2019axial,wang2020axial}. As shown in Figure~\ref{fig:axial-compare}(a), Axial attention proposes to first apply column-wise attention then row-wise, which achieves a global receptive field with $\mathcal{O}(N\sqrt{N})$ complexity (assuming $N$ equals to the number of pixels). On the contrary, our proposed Max-SA shown in Figure~\ref{fig:axial-compare}(b) first employs local attention, then sparse global attention, enjoying global receptive fields with only $\mathcal{O}(N)$ \textit{linear} complexity. Moreover, we deem the proposed Max-SA a more natural approach for vision since the design of attended regions account for the 2D structure of images, \eg mixing tokens in a spatially-local small window.

%%%%% This is a good example for putting figure
%%%%% and table side-by-side
% \begin{figure}
% \begin{minipage}[b]{0.47\textwidth}
% \centering
% \includegraphics[width=0.95\textwidth]{figures/Axial_Max_compare.pdf}
% \caption{This is a Figure by a Table}
% \label{fig:by:table}
% \end{minipage}%
% \hfill%
% \begin{minipage}[b]{0.47\textwidth}
% \centering
% \begin{tabular}{|c|c|} \hline
% Day & Reading \\ \hline\hline
% Monday & 14.6 \\
% Tuesday & 14.3 \\
% Wednesday & 14.2 \\
% Thursday & 14.5 \\
% Friday & 14.9 \\ \hline
% \end{tabular}
% \tabcaption{This is a Table by a Figure}
% \label{table:by:fig}
% \end{minipage}%
% \end{figure}

\subsubsection{MaxViT Block}
We demonstrate in Algo.~\ref{alg:code} an \texttt{einops}-style pseudocode of the MaxViT block which contains MBConv, block attention, and grid attention.

\begin{figure}[!ht]
\vspace{-10mm}
\begin{minipage}{1.\textwidth}
\floatname{algorithm}{Algo.}
\begin{algorithm}[H]
\small
\caption{\small Pseudocode of MaxViT Block}
\label{alg:code}
\begin{lstlisting}[language=python]
# input: features (b, h, w, c). Assume h==w; x/output: features (b, h, w, c).
# p/g: block/grid size. Use 7 by default.

def RelSelfAttn(x): return x # A self-attn function applied on the -2 axis

# Window/grid partition function
from einops import rearrange
def block(x,p):
  return rearrange(x,"b(hy)(wx)c->b(hw)(yx)c",h=x.shape[1]//p,w=x.shape[2]//p,y=p,x=p)
    
def unblock(x,g,p):
  return rearrange(x,"b(hw)(yx)c->b(hy)(wx)c",h=g,w=g,y=p,x=p) 

x = MBConv(input) # MBConv layer

x = block(x,p) # window partition
x = RelSelfAttn(x) # Apply window-attention
x = unblock(x,x.shape[1]//p,p) # reverse

x = block(x,x.shape[1]//g) # grid partition
x = swapaxes(x,-2,-3) # move grid-axis to -2
x = RelSelfAttn(x) # Apply grid-attention
x = swapaxes(x,-2,-3) # reverse swapaxes
output = unblock(x,g,x.shape[1]//g) # reverse
\end{lstlisting}
\end{algorithm}
\end{minipage}
\vspace{-11mm}
\end{figure}

\subsubsection{Classification Head}
Instead of using the \texttt{[cls]} token~\cite{dosovitskiy2020image}, we simply apply global average pooling to the output of the last stage (S4) to obtain the feature representation, followed by the final classification head.

\subsubsection{Architectural Specifications}
Finally, we present detailed architectural specifications for the MaxViT model family (T/S/B/L) in Table~\ref{tab:arch-spec}.

\begin{table}[!t]
\centering
\scriptsize
\setlength{\tabcolsep}{4pt}
\renewcommand{\arraystretch}{1.}
\caption{\textbf{Detailed architectural specifications} for MaxViT families.}
\label{tab:arch-spec}
\begin{tabular}{c|c|c|c}
& \multirow{2}{*}{\begin{tabular}{c}dsp. rate\\(out size)\\\end{tabular}} & \multirow{2}{*}{MaxViT-T} & \multirow{2}{*}{MaxViT-S} \\
& & &  \\\toprule

\multirow{1}{*}{stem} & \begin{tabular}{c}$2\times$\\($112\!\times\!112$)\end{tabular} & 
\begin{tabular}{c}3$\times$3, 64, stride 2\\ 3$\times$3, 64, stride 1\\\end{tabular} &
\begin{tabular}{c}3$\times$3, 64, stride 2\\ 3$\times$3, 64, stride 1\\\end{tabular}
% & \begin{tabular}{c}3$\times$3, 64, stride 2\\ 3$\times$3, 64, stride 1\\\end{tabular}

\\ \hline
\multirow{1}{*}{S1} &
\begin{tabular}{c}$4\times$\\($56\times 56$)\end{tabular} & 
$\left[\begin{array}{c}
\text{MBConv, 64, E 4, R 4} \\
\text{Rel-MSA, P 7$\times$7, H 2} \\
\text{Rel-MSA, G 7$\times$7, H 2} 
\end{array}\right]\times 2 $ &
$\left[\begin{array}{c}
\text{MBConv, 96, E 4, R 4} \\
\text{Rel-MSA, P 7$\times$7, H 3} \\
\text{Rel-MSA, G 7$\times$7, H 3} 
\end{array}\right]\times 2$ 
% & 
% $\left[\begin{array}{c}
% \text{MBConv, 96, E 4, R 4} \\
% \text{Rel-MSA, P 7$\times$7, H 3} \\
% \text{Rel-MSA, G 7$\times$7, H 3} 
% \end{array}\right]\times 2$ 

\\ \hline
\multirow{1}{*}{S2} &
\begin{tabular}{c}$8\times$\\($28\times 28$)\end{tabular} & 
$\left[\begin{array}{c}
\text{MBConv, 128, E 4, R 4} \\
\text{Rel-MSA, P 7$\times$7, H 4} \\
\text{Rel-MSA, G 7$\times$7, H 4} 
\end{array}\right]\times 2$ & 
$\left[\begin{array}{c}
\text{MBConv, 192, E 4, R 4} \\
\text{Rel-MSA, P 7$\times$7, H 6} \\
\text{Rel-MSA, G 7$\times$7, H 6} 
\end{array}\right]\times 2$ 
% & 
% $\left[\begin{array}{c}
% \text{MBConv, 192, E 4, R 4} \\
% \text{Rel-MSA, P 7$\times$7, H 6} \\
% \text{Rel-MSA, G 7$\times$7, H 6} 
% \end{array}\right]\times 6$ 

\\ \hline
\multirow{1}{*}{S3} &
\begin{tabular}{c}$16\times$\\($14\times 14$)\end{tabular} & 
$\left[\begin{array}{c}
\text{MBConv, 256, E 4, R 4} \\
\text{Rel-MSA, P 7$\times$7, H 8} \\
\text{Rel-MSA, G 7$\times$7, H 8} 
\end{array}\right]\times 5$ & 
$\left[\begin{array}{c}
\text{MBConv, 384, E 4, R 4} \\
\text{Rel-MSA, P 7$\times$7, H 12} \\
\text{Rel-MSA, G 7$\times$7, H 12} 
\end{array}\right]\times 5$ 
% & 
% $\left[\begin{array}{c}
% \text{MBConv, 384, E 4, R 4} \\
% \text{Rel-MSA, P 7$\times$7, H 12} \\
% \text{Rel-MSA, G 7$\times$7, H 12} 
% \end{array}\right]\!\times\!14$ 

\\ \hline
\multirow{1}{*}{S4} &
\begin{tabular}{c}$32\times$\\($7\times 7$)\end{tabular} & 
$\left[\begin{array}{c}
\text{MBConv, 512, E 4, R 4} \\
\text{Rel-MSA, P 7$\times$7, H 16} \\
\text{Rel-MSA, G 7$\times$7, H 16} 
\end{array}\right]\times 2$ & 
$\left[\begin{array}{c}
\text{MBConv, 768, E 4, R 4} \\
\text{Rel-MSA, P 7$\times$7, H 24} \\
\text{Rel-MSA, G 7$\times$7, H 24} 
\end{array}\right]\times 2$ 
% & 
% $\left[\begin{array}{c}
% \text{MBConv, 768, E 4, R 4} \\
% \text{Rel-MSA, P 7$\times$7, H 24} \\
% \text{Rel-MSA, G 7$\times$7, H 24} 
% \end{array}\right]\times 2$

\\\bottomrule 
\rule{0pt}{3ex}
% \vspace{2mm}
& \multirow{2}{*}{\begin{tabular}{c}dsp. rate\\(out size)\\\end{tabular}} & \multirow{2}{*}{MaxViT-B} & \multirow{2}{*}{MaxViT-L} \\
& & &  \\\toprule

\multirow{1}{*}{stem} & \begin{tabular}{c}$2\times$\\($112\!\times\!112$)\end{tabular} 
& \begin{tabular}{c}3$\times$3, 64, stride 2\\ 3$\times$3, 64, stride 1\\\end{tabular}
& 
\begin{tabular}{c}3$\times$3, 128, stride 2\\ 3$\times$3, 128, stride 1\\\end{tabular} 
% &
% \begin{tabular}{c}3$\times$3, 192, stride 2\\ 3$\times$3, 192, stride 1\\\end{tabular}  

\\ \cline{1-4}

\multirow{1}{*}{S1} &
\begin{tabular}{c}$4\times$\\($56\times 56$)\end{tabular} 
& 
$\left[\begin{array}{c}
\text{MBConv, 96, E 4, R 4} \\
\text{Rel-MSA, P 7$\times$7, H 3} \\
\text{Rel-MSA, G 7$\times$7, H 3} 
\end{array}\right]\times 2$ 
& 
$\left[\begin{array}{c}
\text{MBConv, 128, E 4, R 4} \\
\text{Rel-MSA, P 7$\times$7, H 4} \\
\text{Rel-MSA, G 7$\times$7, H 4} 
\end{array}\right]\times 2 $ 
% & 
% $\left[\begin{array}{c}
% \text{MBConv, 192, E 4, R 4} \\
% \text{Rel-MSA, P 7$\times$7, H 6} \\
% \text{Rel-MSA, G 7$\times$7, H 6} 
% \end{array}\right]\times 2$

\\ \cline{1-4}

\multirow{1}{*}{S2} &
\begin{tabular}{c}$8\times$\\($28\times 28$)\end{tabular} 
& 
$\left[\begin{array}{c}
\text{MBConv, 192, E 4, R 4} \\
\text{Rel-MSA, P 7$\times$7, H 6} \\
\text{Rel-MSA, G 7$\times$7, H 6} 
\end{array}\right]\times 6$ 
& 
$\left[\begin{array}{c}
\text{MBConv, 256, E 4, R 4} \\
\text{Rel-MSA, P 7$\times$7, H 8} \\
\text{Rel-MSA, G 7$\times$7, H 8} 
\end{array}\right]\times 6$
% & 
% $\left[\begin{array}{c}
% \text{MBConv, 384, E 4, R 4} \\
% \text{Rel-MSA, P 7$\times$7, H 12} \\
% \text{Rel-MSA, G 7$\times$7, H 12} 
% \end{array}\right]\times 6$ 

\\ \cline{1-4}

\multirow{1}{*}{S3} &
\begin{tabular}{c}$16\times$\\($14\times 14$)\end{tabular}
& 
$\left[\begin{array}{c}
\text{MBConv, 384, E 4, R 4} \\
\text{Rel-MSA, P 7$\times$7, H 12} \\
\text{Rel-MSA, G 7$\times$7, H 12} 
\end{array}\right]\!\times\!14$ 
& 
$\left[\begin{array}{c}
\text{MBConv, 512, E 4, R 4} \\
\text{Rel-MSA, P 7$\times$7, H 16} \\
\text{Rel-MSA, G 7$\times$7, H 16} 
\end{array}\right]\times 14$ 
% & 
% $\left[\begin{array}{c}
% \text{MBConv, 768, E 4, R 4} \\
% \text{Rel-MSA, P 7$\times$7, H 24} \\
% \text{Rel-MSA, G 7$\times$7, H 24} 
% \end{array}\right]\times 14$

\\ \cline{1-4}

\multirow{1}{*}{S4} &
\begin{tabular}{c}$32\times$\\($7\times 7$)\end{tabular}
& 
$\left[\begin{array}{c}
\text{MBConv, 768, E 4, R 4} \\
\text{Rel-MSA, P 7$\times$7, H 24} \\
\text{Rel-MSA, G 7$\times$7, H 24} 
\end{array}\right]\times 2$
& 
$\left[\begin{array}{c}
\text{MBConv, 1024, E 4, R 4} \\
\text{Rel-MSA, P 7$\times$7, H 32} \\
\text{Rel-MSA, G 7$\times$7, H 32} 
\end{array}\right]\times 2$ 
% & 
% $\left[\begin{array}{c}
% \text{MBConv, 1536, E 4, R 4} \\
% \text{Rel-MSA, P 7$\times$7, H 48} \\
% \text{Rel-MSA, G 7$\times$7, H 48} 
% \end{array}\right]\times 2$ \\

\end{tabular}
\vspace{-1mm}
\end{table}

\subsection{Detection and Segmentation Models}
\label{ssec:detection-model}
We follow the settings of the cascaded Faster-RCNN~\cite{ren2015fasterrcnn} and Mask-RCNN~\cite{he2017maskrcnn}, but replace the feature extraction backbone with our MaxViT backbone.
We also applied FPN~\cite{Lin2017FeaturePN} in the feature map generation, where the S2, S3, S4 (multi-scale features of targeted resolution $1/8$, $1/16$, $1/32$ in MaxViT, respectively) are used.
Then the generated feature maps are fed into the detection head.
For fair comparison, we follow the original implementation without adopting any system-level strategies to further boost the final performance, such as the HTC framework~\cite{chen2019hybrid}, instaboost~\cite{fang2019instaboost}, \etc used in Swin~\cite{liu2021swin}.
We show the results of MaxViT-T/S/B on these two tasks to compare it against recent strong models at similar model complexity.

\subsection{Image Aesthetics Model}
\label{ssec:image-quality-model}

This task requires incorporating both local and global information of an image to accurately predict human perceptual preference. To this end, the model needs to have the capacity to learn pixel-level quality aspects such as sharpness, noisiness and contrast as well as semantic-level aspects such as composition and depth-of-field. We follow~\cite{talebi2018nima} and use the normalized Earth Mover's Distance as our training loss. Given the ground truth and predicted probability mass functions $\textbf{p}$ and $\widehat{\textbf{p}}$ representing the histogram of scores, the normalized Earth Mover's Distance can be expressed as:
\begin{equation}
\label{eqn:emd}
\mbox{EMD}(\textbf{p}, \widehat{\textbf{p}}) = \left( \frac{1}{N} \sum_{k=1}^{N} |\mbox{CDF}_{\textbf{p}}(k) - \mbox{CDF}_{\widehat{\textbf{p}}}(k)|^r \right)^{1/r}
\end{equation}
where $\mbox{CDF}_{\textbf{p}}(k)$ is the cumulative distribution function as $\sum_{i=1}^{k} \textbf{p}_{i}$, and $N=10$ represents the number score bins. In our experiments we set $r=2$. We remove the classification head used in MaxViT, and instead append a fully-connected layer with 10 neurons followed by $\mathsf{softmax}$.

% We train and evaluate the MaxViT model on the AVA benchmark~\cite{murray2012ava}. This dataset consists of ~255,000 images rated by armature photographers through photography contests. Each image is rated by an average of 200 human raters, assigning a score from 1 to 10 to images. The higher the score, the better the visual quality of the image. Each image in the dataset has a histogram of scores associated with it, which we use as the ground truth label. 
% \subsubsection{Results on AVA.} Similar to \cite{talebi2018nima}, we split the dataset into train and test sets, such that 20\% of the data is used for testing. We train MaxViT for three different input resolutions: $224 \times 224$, $384 \times 384$ and $512 \times 512$. Also, as it is a common practice in quality assessment, we initialized the model with ImageNet weights. To evaluate and compare our model with existing methods, we present a summary of our results in Table~\ref{tab:iqa-comparison}. For similar input resolutions, the proposed MaxViT model outperforms existing image quality assessment methods. As the input resolution increases, the performance improves. Also, MaxViT shows better linear correlation compared to the state-of-the-art method~\cite{ke2021musiq} which uses multi- and full-resolution inputs. 

% \textcolor{red}{@htalebi}

\begin{figure}[!t]
\centering
\includegraphics[width=0.94\linewidth]{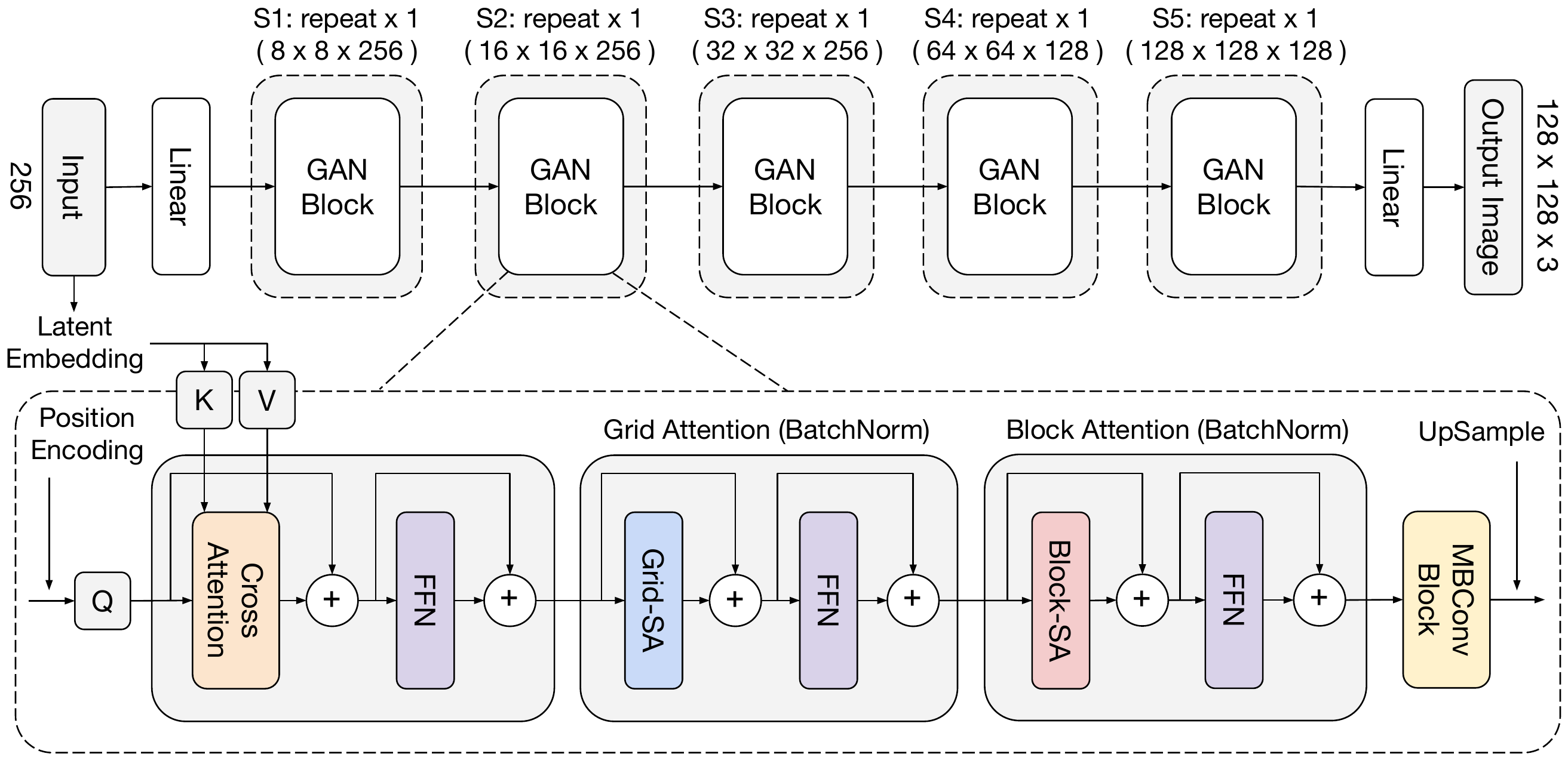}
\caption{\textbf{Generator architecture using the MaxViT block for the GAN experiment.}
In every stage, we first use the cross-attention module to let the features attend to the latent embedding projected from the input code, which are then fed into the proposed MaxViT block consisting of grid attention, block attention, and MBConv layer. Note that unlike the main model in Sec.~\ref{ssec:main-model}, the order of applying the three layers are reversed: from global to local.}
\label{fig:gan-architecture}
\end{figure}

\subsection{GAN Model}
\label{ssec:gan-model}
The above image recognition tasks can validate the power of our proposed MaxViT block used in downsampling (contracting) models.
For this GAN experiment, we would like to demonstrate its effectiveness in upsampling (expanding) architectures.
The MaxViT-GAN model for image generation is illustrated in Figure~\ref{fig:gan-architecture}.
For unconditional image generation, MaxViT-GAN first takes a latent code $z\sim \mathcal{N}(\mathbf{0},\mathbf{I})$ as input, then progressively generates an image of target resolution through a hierarchically upsampling structure. We start by linearly projecting the input to a feature with spatial dimension $8\times 8$. During the generation, the feature will go through five stages consisting of identical GAN blocks with gradually increased spatial resolution, similar to the design of our main model. Similar to~\cite{zhao2021improved}, we apply a cross-attention layer before the MaxViT block as a memory-efficient form of self-modulation in every stage, which has been shown to stabilize GAN training and also improve mode coverage~\cite{chen2018self,zhao2021improved}. We use pixel shuffle~\cite{shi2016real} for upsampling in the end of each stage.

% \textcolor{red}{@zhanghan}

\section{Experimental Settings}
\label{sec:experimental-settings}

\subsection{ImageNet Classification}
\label{ssec:imagenet-settings}

We provide ImageNet-1K experimental settings of MaxViT models for both pre-training and fine-tuning in Table~\ref{tab:i1k-experimental-settings}. All the MaxViT variants used similar hyperparameters except that we mainly customize the stochastic depth rate to regularize each model separately.

\begin{table}[!ht]
\scriptsize
\centering
\setlength{\tabcolsep}{2pt}
\renewcommand{\arraystretch}{1.1}
\caption{\textbf{Detailed hyperparameters used in ImageNet-1K experiments.} Multiple values separated by `$/$' are for each model size respectively.}
\label{tab:i1k-experimental-settings}
\begin{tabular}{l|cc|cc|cc}
% \toprule
\multirow{3}{*}{Hyperparameter}  & \multicolumn{2}{c|}{\textbf{ImageNet-1K}} & \multicolumn{2}{c|}{\textbf{ImageNet-21K}} &
\multicolumn{2}{c}{\textbf{JFT-300M}} \\
& Pre-training & Fine-tuning & Pre-training & Fine-tuning & Pre-training & Fine-tuning \\
& \multicolumn{2}{c|}{(MaxViT-T/S/B/L)}
& \multicolumn{2}{c|}{(MaxViT-B/L/XL)}
& \multicolumn{2}{c}{(MaxViT-B/L/XL)} \\
\toprule
Stochastic depth & \multicolumn{2}{c|}{$0.2/0.3/0.4/0.6$} & \multicolumn{1}{c}{$0.3/0.4/0.6$} &
\multicolumn{1}{c|}{$0.4/0.5/0.9$} &
\multicolumn{1}{c}{$0.0/0.0/0.0$} & \multicolumn{1}{c}{$0.1/0.2/0.2$} \\
Center crop  & True & False & True & False & True & False \\
RandAugment & 2, 15 & 2, 15 & 2, 5 & 2, 15 & 2, 5 & 2, 15 \\
Mixup alpha & 0.8 & 0.8 & None & None & None & None  \\
Loss type & Softmax & Softmax & Sigmoid & Softmax & Sigmoid & Softmax \\
Label smoothing & 0.1 & 0.1 & 0.0001 & 0.1 & 0 & 0.1 \\
Train epochs & 300 & 30 & 90 & 30 & 14 & 30 \\
Train batch size & 4096 & 512 & 4096 & 512 & 4096 & 512 \\
Optimizer type & AdamW & AdamW & AdamW & AdamW & AdamW & AdamW \\
Peak learning rate & 3e-3 & 5e-5 & 1e-3 & 5e-5 & 1e-3 & 5e-5 \\
Min learning rate & 1e-5 & 5e-5 & 1e-5 & 5e-5 & 1e-5 & 5e-5 \\
Warm-up & 10K steps & None & 5 epochs & None & 20K steps & None \\
LR decay schedule & Cosine & None & Linear & None & Linear & None \\
Weight decay rate & 0.05 & 1e-8 & 0.01 & 1e-8 & 0.01 & 1e-8 \\
Gradient clip & 1.0 & 1.0 & 1.0 & 1.0 & 1.0 & 1.0 \\
EMA decay rate & None & 0.9999 & None & 0.9999 & None & 0.9999 \\
% \bottomrule
\end{tabular}
\end{table}

\subsection{Coco Detection and Segmentation}
\label{ssec:coco-settings}
We evaluated MaxViT on the COCO2017~\cite{lin2014microsoft} object bounding box detection and instance segmentation tasks.
The dataset contains 118K training and 5K validation samples.
All the MaxViT backbones used are pretrained on ImageNet-1k at  resolution $224\times224$.
These pretrained checkpoints are then used as the warm-up weights for fine-tuning the detection and segmentation tasks.
For both tasks, the input images are resized to $896\times896$.
The training is conducted with a batch size of 256, using the AdamW~\cite{loshchilov2017decoupled} optimizer with learning rate of 1e-3, 3e-3, 3e-3, and stochastic depth of $0.8, 0.3, 0.3$ for MaxViT-T/S/B, respectively.

\subsection{Image Aesthetics Assessment}
\label{ssec:image-quality-settings}

We trained and evaluated the MaxViT model on the AVA benchmark~\cite{murray2012ava}. This dataset consists of 255K images rated by armature photographers through photography contests. Each image is rated by an average of 200 human raters, assigning a score from 1 to 10 to images. The higher the score, the better the visual aesthetic quality of the image. Each image in the dataset has a histogram of scores associated with it, which we use as the ground truth label. Similar to \cite{talebi2018nima,ke2021musiq}, we split the dataset into train and test sets, such that 20\% of the data is used for testing. We train MaxViT for three different input resolutions: $224 \times 224$, $384 \times 384$ and $512 \times 512$. We initialized the model with ImageNet-1K 224$\times$224 pre-trained weights. The weight and bias momentums
are set to 0.9, and a dropout rate of 0.75 is applied on the
last layer of the baseline network. We use an initial learning rate of 1e-3, exponentially decayed with decay factor 0.9 every 10 epochs. We set the stochastic depth rate to 0.5.

% \textcolor{red}{@htalebi}

\subsection{Image Generation}
\label{ssec:gan-settings}

We use a ResNet-based discriminator following~\cite{karras2020analyzing}.
To train the model, we also used the standard non-saturating logistic GAN loss with $R1$ gradient penalty~\cite{mescheder2018training} applied to the discriminator with the gradient penalty weight set to 10.
We employ the Adam~\cite{kingma2014adam} optimizer with a learning rate of 1e-4 for both generator and discriminator.
The model is trained on TPU for one million steps with batch size 256.
Notably, we do not employ extra GAN training tricks such as pixel norm, noise injection, progressive growing, \etc on which recent state-of-the-art models are heavily relied to attain good results~\cite{karras2017progressive,karras2020analyzing}.
The overall objectives of the GAN training are defined as:
\begin{align}
\label{eq:objective-gan}
\mathcal{L}_G& = -\mathbb{E}_{z\sim P_z}[\log(D(G(z))],\\
\mathcal{L}_D& = -\mathbb{E}_{x\sim P_x}[\log(D(x))]-\mathbb{E}_{z\sim P_z}[\log(1-D(G(z)))]+\gamma\mathbb{E}_{x\sim P_x}[\|\nabla_xD(x) \|_2^2],
\end{align}
where $\gamma$ denotes the $R_1$ gradient penalty weight.

% \textcolor{red}{@zhanghan}

\section{Complete Experimental Results}
\label{sec:complete-experimental-results}

We provide complete experiment comparisons for ImageNet-1K, Image-21K, and JFT datasets in Table~\ref{tab:imagenet-1k-complete} and Table~\ref{tab:imagenet21k-jft-complete}, respectively. We also provide more visual results for unconditional image generation on ImageNet-1K in Figure~\ref{fig:more-gan-results}.

\begin{table}[!t]
\vspace{-4mm}
\centering
\scriptsize
\setlength{\tabcolsep}{4pt}
\renewcommand{\arraystretch}{1.}
\caption{Complete performance comparison under ImageNet-1K only setting.}
\label{tab:imagenet-1k-complete}
\begin{tabular}{c|l|ccccc}
% \hline
% \rowcolor[gray]{0.95}
& \multirow{2}{*}{Model} & \multirow{2}{*}{\begin{tabular}{@{}c@{}}Eval\\size\end{tabular}} & \multirow{2}{*}{Params} & \multirow{2}{*}{FLOPs} &
\multirow{2}{*}{\begin{tabular}{@{}c@{}}throughput\\ (img/s)\\ \end{tabular}} &
\multirow{2}{*}{\begin{tabular}{@{}c@{}}ImageNet \\top-1 acc.\\ \end{tabular}} \\
& & & & & \\ 
% \rowcolor[gray]{0.95}
\toprule
\multirow{20}{*}{ConvNets} &
\textcolor{blueish}{$\bullet$}EffNet-B3~\cite{tan2019efficientnet} & 300  & 12M & 1.8G & 732.1 & 81.6 \\
& \textcolor{blueish}{$\bullet$}EffNet-B4~\cite{tan2019efficientnet} & 380  & 19M & 4.2G & 349.4 & 82.9 \\
& \textcolor{blueish}{$\bullet$}EffNet-B5~\cite{tan2019efficientnet} & 456  & 30M & 9.9G & 169.1 & 83.6 \\
& \textcolor{blueish}{$\bullet$}EffNet-B6~\cite{tan2019efficientnet} & 528  & 43M & 19.0G & 96.9 & 84.0 \\
& \textcolor{blueish}{$\bullet$}EffNet-B7~\cite{tan2019efficientnet} & 600  & 66M & 37.0G & 55.1 & 84.3 \\
% \textcolor{blueish}{$\bullet$}RegNetY-4GF~\cite{radosavovic2020designing} & 224 & 21M &  4.0G & 1156.7 & 80.0 \\
& \textcolor{blueish}{$\bullet$}RegNetY-8GF~\cite{radosavovic2020designing} & 224 & 39M & 8.0G & 591.6 & 81.7 \\
& \textcolor{blueish}{$\bullet$}RegNetY-16GF~\cite{radosavovic2020designing} & 224 & 84M & 16.0G & 334.7 & 82.9 \\
& \textcolor{blueish}{$\bullet$}NFNet-F0~\cite{brock2021high} & 256 & 72M & 12.4G & 533,.3 & 83.6 \\
& \textcolor{blueish}{$\bullet$}NFNet-F1~\cite{brock2021high} & 320 & 132M & 35.5G & 228.5 & 84.7 \\
& \textcolor{blueish}{$\bullet$}NFNet-F2~\cite{brock2021high} & 352 & 194M & 62.6G & 129.0 & 85.1  \\
& \textcolor{blueish}{$\bullet$}NFNet-F3~\cite{brock2021high} & 416 & 255M & 114.7G & 78.8 & 85.7 \\
& \textcolor{blueish}{$\bullet$}NFNet-F4~\cite{brock2021high} & 512 & 316M & 215.2G & 51.7 &  85.9 \\
& \textcolor{blueish}{$\bullet$}NFNet-F5~\cite{brock2021high} & 544 & 377M & 289.8G & - & 86.0 \\
& \textcolor{blueish}{$\bullet$}EffNetV2-S~\cite{tan2021efficientnetv2} & 384 & 24M & 8.8G & 666.6 & 83.9 \\
& \textcolor{blueish}{$\bullet$}EffNetV2-M~\cite{tan2021efficientnetv2} & 380 & 55M & 24.0G & 280.7 & 85.1 \\
& \textcolor{blueish}{$\bullet$}EffNetV2-L~\cite{tan2021efficientnetv2} & 480 & 121M & 53.0G & 163.2 & 85.7 \\
& \textcolor{blueish}{$\bullet$}ConvNeXt-T~\cite{liu2022convnet} & 224 & 29M & 4.5G & 774.7 & 82.1 \\
& \textcolor{blueish}{$\bullet$}ConvNeXt-S~\cite{liu2022convnet} & 224 & 50M & 8.7G & 447.1 & 83.1 \\
& \textcolor{blueish}{$\bullet$}ConvNeXt-B~\cite{liu2022convnet} & 224 & 89M & 15.4G & 292.1 & 83.8 \\
% \textcolor{blueish}{$\bullet$}ConvNeXt-L~\cite{liu2022convnet} & 224 & 198M & 34.4G & 146.8 & 84.3 \\
& \textcolor{blueish}{$\bullet$}ConvNeXt-L~\cite{liu2022convnet} & 384 & 198M & 101.0G & 50.4 & 85.5 \\
\midrule

\multirow{20}{*}{ViTs} &
\textcolor{brickred}{$\circ$}ViT-B/32~\cite{dosovitskiy2020image} & 384 & 86M & 55.4G & 85.9 & 77.9 \\
& \textcolor{brickred}{$\circ$}ViT-B/16~\cite{dosovitskiy2020image} & 384 & 307M & 190.7G & 27.3 & 76.5 \\
& \textcolor{brickred}{$\circ$}DeiT-S~\cite{touvron2021training} & 224 & 22M & 4.6G & 940.4 & 79.8 \\
& \textcolor{brickred}{$\circ$}DeiT-B~\cite{touvron2021training} & 224 & 86M & 17.5G & 292.3 & 81.8 \\
& \textcolor{brickred}{$\circ$}DeiT-B~\cite{touvron2021training} & 384 & 86M & 55.4G & 85.9 & 83.1 \\
& \textcolor{brickred}{$\circ$}CaiT-S36~\cite{touvron2021going} & 224 & 68M & 13.9G & - & 83.3 \\
& \textcolor{brickred}{$\circ$}CaiT-M24~\cite{touvron2021going} & 224 & 186M & 36.0G & - & 83.4 \\
& \textcolor{brickred}{$\circ$}CaiT-M24~\cite{touvron2021going} & 384 & 186M & 116.1G & - & 84.5 \\
& \textcolor{brickred}{$\circ$}DeepViT-S~\cite{zhou2021deepvit} & 224 & 27M & 6.2G & - & 82.3 \\
& \textcolor{brickred}{$\circ$}DeepViT-L~\cite{zhou2021deepvit} & 224 & 55M & 12.5G & - & 83.1 \\
& \textcolor{brickred}{$\circ$}T2T-ViT-14~\cite{yuan2021tokens} & 224 & 22M & 6.1G & - & 81.7 \\
& \textcolor{brickred}{$\circ$}T2T-ViT-19~\cite{yuan2021tokens} & 224 & 39M & 9.8G & - & 82.2 \\
& \textcolor{brickred}{$\circ$}T2T-ViT-24~\cite{yuan2021tokens} & 224 & 64M & 15.0G & - & 82.6 \\
& \textcolor{brickred}{$\circ$}Swin-T~\cite{liu2021swin} & 224 & 29M & 4.5G & 755.2 & 81.3 \\
& \textcolor{brickred}{$\circ$}Swin-S~\cite{liu2021swin} & 224 & 50M & 8.7G & 436.9 & 83.0 \\
% \textcolor{brickred}{$\circ$}Swin-B~\cite{liu2021swin} & 224 & 88M & 15.4G & 278.1 & 83.5 \\
& \textcolor{brickred}{$\circ$}Swin-B~\cite{liu2021swin} & 384 & 88M & 47.0G & 84.7 & 84.5 \\
& \textcolor{brickred}{$\circ$}CSwin-B~\cite{dong2021cswin} & 224 & 78M & 15.0G & 250 & 84.2 \\
& \textcolor{brickred}{$\circ$}CSwin-B~\cite{dong2021cswin} & 384 & 78M & 47.0G & - & 85.4 \\
& \textcolor{brickred}{$\circ$}Focal-S~\cite{yang2021focal} & 224 & 51M & 9.1G & - & 83.5 \\
& \textcolor{brickred}{$\circ$}Focal-B~\cite{yang2021focal} & 224 & 90M & 16.0G & - & 83.8 \\
\midrule

\multirow{17}{*}{Hybrid} &
\textcolor{darkgreen}{$\diamond$}CvT-13~\cite{wu2021cvt} & 224 & 20M & 4.5G & - & 81.6 \\
& \textcolor{darkgreen}{$\diamond$}CvT-21~\cite{wu2021cvt} & 224 & 32M & 7.1G & - & 82.5 \\
& \textcolor{darkgreen}{$\diamond$}CvT-21~\cite{wu2021cvt} & 384 & 32M & 24.9G & - & 83.3 \\
& \textcolor{darkgreen}{$\diamond$}CoAtNet-0~\cite{dai2021coatnet} & 224 & 25M & 4.2G & 534.5 & 81.6 \\
& \textcolor{darkgreen}{$\diamond$}CoAtNet-1~\cite{dai2021coatnet} & 224 & 42M & 8.4G & 336.5 & 83.3 \\
& \textcolor{darkgreen}{$\diamond$}CoAtNet-2~\cite{dai2021coatnet} & 224 & 75M & 15.7G & 247.6 &  84.1 \\
% & \textcolor{darkgreen}{$\diamond$}CoAtNet-3~\cite{dai2021coatnet} & 224 & 168M & 34.7G & 163.3 & 84.5 \\
& \textcolor{darkgreen}{$\diamond$}CoAtNet-3~\cite{dai2021coatnet} & 384 & 168M & 107.4G & 48.5 & 85.8 \\
& \textcolor{darkgreen}{$\diamond$}CoAtNet-3~\cite{dai2021coatnet} & 512 & 168M & 203.1G & 22.4 & 86.0 \\
\cmidrule(lr){2-7}
& \textcolor{darkgreen}{$\diamond$}MaxViT-T  & 224 & 31M & 5.6G & 349.6 & 83.62 \\
& \textcolor{darkgreen}{$\diamond$}MaxViT-S  & 224 & 69M & 11.7G & 242.5 & 84.45 \\
& \textcolor{darkgreen}{$\diamond$}MaxViT-B & 224 & 120M & 23.4G & 133.6 & 84.95 \\
& \textcolor{darkgreen}{$\diamond$}MaxViT-L  & 224 &  212M & 43.9G & 99.4 & 85.17 \\
\cmidrule(lr){2-7}
& \textcolor{darkgreen}{$\diamond$}MaxViT-T  & 384 & 31M & 17.7G & 121.9 & 85.24 \\
& \textcolor{darkgreen}{$\diamond$}MaxViT-S & 384 & 69M & 36.1G & 82.7 & 85.74 \\
& \textcolor{darkgreen}{$\diamond$}MaxViT-B & 384 & 120M & 74.2G & 45.8 & 86.34 \\
& \textcolor{darkgreen}{$\diamond$}MaxViT-L & 384 &  212M & 133.1G & 34.3 & 84.40 \\
\cmidrule(lr){2-7}
& \textcolor{darkgreen}{$\diamond$}MaxViT-T & 512 & 31M & 33.7G &  63.8 & 85.72 \\
& \textcolor{darkgreen}{$\diamond$}MaxViT-S & 512 & 69M & 67.6G & 43.3 & 86.19 \\
& \textcolor{darkgreen}{$\diamond$}MaxViT-B & 512 & 120M & 138.5G & 24.0 & 86.66 \\
& \textcolor{darkgreen}{$\diamond$}MaxViT-L & 512 &  212M & 245.4G & 17.8 & \textbf{86.70} \\
\bottomrule
\end{tabular}
\end{table}

\begin{table}[!ht]
\centering
\scriptsize
\setlength{\tabcolsep}{4pt}
\renewcommand{\arraystretch}{1.}
\caption{Complete performance comparison for ImageNet-21K and JFT pre-trained models.}
\label{tab:imagenet21k-jft-complete}
\begin{tabular}{c|l|ccccc}
& \multirow{2}{*}{Model} & \multirow{2}{*}{\begin{tabular}{@{}c@{}}Eval\\size\end{tabular}} & \multirow{2}{*}{Params} & \multirow{2}{*}{FLOPs} &
\multicolumn{2}{c}{IN-1K top-1 acc.} \\\cline{6-7}
% \vspace{0.5mm}
\rule{0pt}{2ex}  & & & & & 21K$\rightarrow$1K & JFT$\rightarrow$1K\\ 
\toprule
\multirow{10}{*}{ConvNets} &
\textcolor{blueish}{$\bullet$}BiT-R-101x3~\cite{kolesnikov2020big} & 384 & 388M & 204.6G  & 84.4 \\
& \textcolor{blueish}{$\bullet$}BiT-R-152x4~\cite{kolesnikov2020big} & 480 & 937M & 840.5G  & 85.4 \\
% \textcolor{blueish}{$\bullet$}EffNet-B7~\cite{tan2019efficientnet} & \\
&\textcolor{blueish}{$\bullet$}EffNetV2-S~\cite{tan2021efficientnetv2} & 384 & 24M & 8.8G  & 85.0 \\
&\textcolor{blueish}{$\bullet$}EffNetV2-M~\cite{tan2021efficientnetv2} & 480 & 55M & 24.0G  & 86.1 \\
&\textcolor{blueish}{$\bullet$}EffNetV2-L~\cite{tan2021efficientnetv2} & 480 & 121M & 53.0G  & 86.8 \\
&\textcolor{blueish}{$\bullet$}EffNetV2-XL~\cite{tan2021efficientnetv2} & 512 & 208M & 94.0G  & 87.3 \\
& \textcolor{blueish}{$\bullet$}NFNet-F4+~\cite{brock2021high} & 512 & 527M & 367G & - & 89.20  \\
&\textcolor{blueish}{$\bullet$}ConvNeXt-B~\cite{liu2022convnet} & 384 & 89M & 45.1G &86.8 \\
&\textcolor{blueish}{$\bullet$}ConvNeXt-L~\cite{liu2022convnet} & 384 & 198M & 101.0G & 87.5 \\
&\textcolor{blueish}{$\bullet$}ConvNeXt-XL~\cite{liu2022convnet} & 384 & 350M & 179.0G & 87.8 \\
\midrule
\multirow{12}{*}{ViTs} &
\textcolor{brickred}{$\circ$}ViT-B/16~\cite{dosovitskiy2020image} & 384 & 87M & 55.5G  & 84.0 \\
&\textcolor{brickred}{$\circ$}ViT-L/16~\cite{dosovitskiy2020image} & 384 & 305M & 191.1G & 85.2 \\
& \textcolor{brickred}{$\circ$}ViT-L/16~\cite{dosovitskiy2020image} & 512 & 305M & 364G &  - & 87.76 \\
& \textcolor{brickred}{$\circ$}ViT-H/14~\cite{dosovitskiy2020image} & 518 & 632M & 1021G &  - & 88.55 \\
&\textcolor{brickred}{$\circ$}HaloNet-H4~\cite{vaswani2021scaling} & 384 & 85M & - & 85.6  \\
&\textcolor{brickred}{$\circ$}HaloNet-H4~\cite{vaswani2021scaling} & 512 & 85M & - & 85.8 \\
&\textcolor{brickred}{$\circ$}Swin-B~\cite{liu2021swin} & 384 & 88M & 47.0G &  86.4 \\
&\textcolor{brickred}{$\circ$}Swin-L~\cite{liu2021swin} & 384 & 197M & 103.9G & 87.3 \\
&\textcolor{brickred}{$\circ$}SwinV2-B~\cite{liu2021swin} & 384 & 88M & - &  87.1 \\
&\textcolor{brickred}{$\circ$}SwinV2-L~\cite{liu2021swin} & 384 & 197M & - &  87.7 \\
&\textcolor{brickred}{$\circ$}CSwin-B~\cite{dong2021cswin} & 384 & 78M & 47.0G &  87.0 \\
&\textcolor{brickred}{$\circ$}CSwin-L~\cite{dong2021cswin} & 384 & 173M & 96.8G &  87.5 \\
\midrule
\multirow{17}{*}{Hybrid} &
\textcolor{darkgreen}{$\diamond$}CvT-13~\cite{wu2021cvt} & 384 & 20M & 16.0G &  83.3 \\
&\textcolor{darkgreen}{$\diamond$}CvT-21~\cite{wu2021cvt} & 384 & 32M & 25.0G &  84.9 \\
&\textcolor{darkgreen}{$\diamond$}CvT-W24~\cite{wu2021cvt} & 384 & 277M & 193.2G &  87.7 \\
&\textcolor{darkgreen}{$\diamond$}ResNet+ViT-L/16~\cite{dosovitskiy2020image} & 384 & 330M & - & - & 87.12 \\
&\textcolor{darkgreen}{$\diamond$}CoAtNet-2~\cite{dai2021coatnet} & 384 & 75M & 49.8G &  87.1 \\
&\textcolor{darkgreen}{$\diamond$}CoAtNet-3~\cite{dai2021coatnet} & 384 & 168M & 107.4G &  87.6 \\
&\textcolor{darkgreen}{$\diamond$}CoAtNet-4~\cite{dai2021coatnet} & 384 & 275M & 189.5G &  87.9 \\
&\textcolor{darkgreen}{$\diamond$}CoAtNet-2~\cite{dai2021coatnet} & 512 & 75M & 96.7G &  87.3 \\
&\textcolor{darkgreen}{$\diamond$}CoAtNet-3~\cite{dai2021coatnet} & 512 & 168M & 203.1G &  87.9 & 88.81 \\
&\textcolor{darkgreen}{$\diamond$}CoAtNet-4~\cite{dai2021coatnet} & 512 & 275M & 360.9G &  88.1 & 89.11 \\
&\textcolor{darkgreen}{$\diamond$}CoAtNet-5~\cite{dai2021coatnet} & 512 & 688M & 812G & -  & \textbf{89.77} \\
\cmidrule(lr){2-7}
& \textcolor{darkgreen}{$\diamond$}MaxViT-B & 384 & 119M & 74.2G  & 88.24 & 88.69 \\
& \textcolor{darkgreen}{$\diamond$}MaxViT-L & 384 & 212M & 128.7G & 88.32 & 89.12 \\
& \textcolor{darkgreen}{$\diamond$}MaxViT-XL & 384 & 475M & 293.7G & 88.51 &  89.36 \\
\cmidrule(lr){2-7}
& \textcolor{darkgreen}{$\diamond$}MaxViT-B & 512 & 119M & 138.3G & 88.38 &  88.82 \\
& \textcolor{darkgreen}{$\diamond$}MaxViT-L & 512 &  212M & 245.2G & 88.46 &  89.41 \\
& \textcolor{darkgreen}{$\diamond$}MaxViT-XL & 512 & 475M & 535.2G & \textbf{88.70} & {89.53}  \\
% MaxViT-XL throughput @ 224: 65.97
\bottomrule
% \multicolumn{6}{c}{\rule{0pt}{3ex}    \textbf{(b) JFT-300M pre-trained models}} \\
% \toprule
% \multirow{2}{*}{Model} & \multirow{2}{*}{\begin{tabular}{@{}c@{}}Eval\\size\\\end{tabular}} & \multirow{2}{*}{Params} & \multirow{2}{*}{FLOPs} &
% \multirow{2}{*}{\begin{tabular}{@{}c@{}}TPUv3-\\ core-days\\ \end{tabular}} &
% \multirow{2}{*}{\begin{tabular}{@{}c@{}}ImageNet\\top-1 acc.\\ \end{tabular}} \\
% \\
% \midrule
% \textcolor{darkgreen}{$\diamond$}ResNet+ViT-L/16~\cite{dosovitskiy2020image} & 384 & 330M & - & - & 87.12 \\
% \textcolor{brickred}{$\circ$}ViT-L/16~\cite{dosovitskiy2020image} & 512 & 307M & 364G & 0.68K & 87.76 \\
% \textcolor{brickred}{$\circ$}ViT-H/14~\cite{dosovitskiy2020image} & 518 & 632M & 1021G & 2.5K & 88.55 \\
% \textcolor{blueish}{$\bullet$}NFNet-F4+~\cite{brock2021high} & 512 & 527M & 367G & 1.86K & 89.2 \\
% \textcolor{darkgreen}{$\diamond$}CoAtNet-3~\cite{dai2021coatnet} & 512 & 168M & 214G & 0.58K & 88.81 \\
% \textcolor{darkgreen}{$\diamond$}CoAtNet-4~\cite{dai2021coatnet} & 512 & 275M & 361G & 0.95K & 89.11 \\
% \textcolor{darkgreen}{$\diamond$}CoAtNet-5~\cite{dai2021coatnet} & 512 & 688M & 812G &  1.82K & 89.77 \\
% \midrule
% \textcolor{darkgreen}{$\diamond$}MaxViT-B & 512 \\
% \textcolor{darkgreen}{$\diamond$}MaxViT-L & 512 \\
% \textcolor{darkgreen}{$\diamond$}MaxViT-XL & 512 \\
% \bottomrule
\end{tabular}
\end{table}

\begin{figure}[!h]
\def\xwidth{0.166}%{0.135}
\def\xem{-3.55pt}
\setlength{\tabcolsep}{0pt}
\centering
% \renewcommand\thefigure{F1}
% \vspace{-0.3cm}
\begin{tabular}{@{}cccccc@{}}
\includegraphics[width=\xwidth\linewidth]{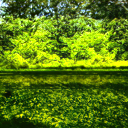} &
\includegraphics[width=\xwidth\linewidth]{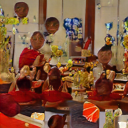} &
\includegraphics[width=\xwidth\linewidth]{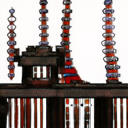} &
\includegraphics[width=\xwidth\linewidth]{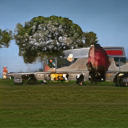} &
\includegraphics[width=\xwidth\linewidth]{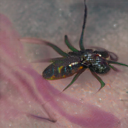} &
\includegraphics[width=\xwidth\linewidth]{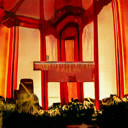}\\[\xem]
\includegraphics[width=\xwidth\linewidth]{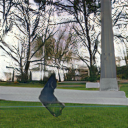} &
\includegraphics[width=\xwidth\linewidth]{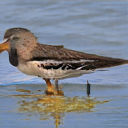} &
\includegraphics[width=\xwidth\linewidth]{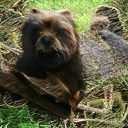} &
\includegraphics[width=\xwidth\linewidth]{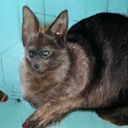} &
\includegraphics[width=\xwidth\linewidth]{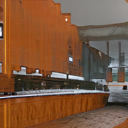} &
\includegraphics[width=\xwidth\linewidth]{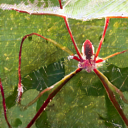}\\[\xem]
\includegraphics[width=\xwidth\linewidth]{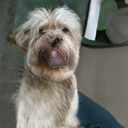} &
\includegraphics[width=\xwidth\linewidth]{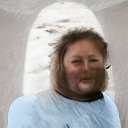} &
\includegraphics[width=\xwidth\linewidth]{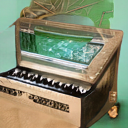} &
\includegraphics[width=\xwidth\linewidth]{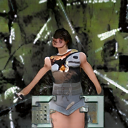} &
\includegraphics[width=\xwidth\linewidth]{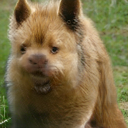} &
\includegraphics[width=\xwidth\linewidth]{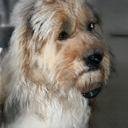}\\[\xem]
\includegraphics[width=\xwidth\linewidth]{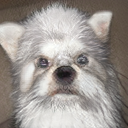} &
\includegraphics[width=\xwidth\linewidth]{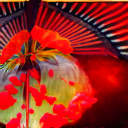} &
\includegraphics[width=\xwidth\linewidth]{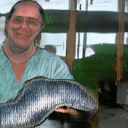} &
\includegraphics[width=\xwidth\linewidth]{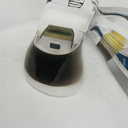} &
\includegraphics[width=\xwidth\linewidth]{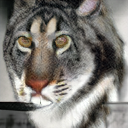} &
\includegraphics[width=\xwidth\linewidth]{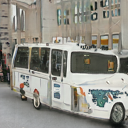}\\[\xem]
\includegraphics[width=\xwidth\linewidth]{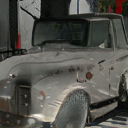} &
\includegraphics[width=\xwidth\linewidth]{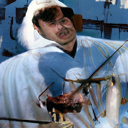} &
\includegraphics[width=\xwidth\linewidth]{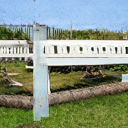} &
\includegraphics[width=\xwidth\linewidth]{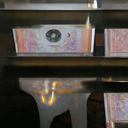} &
\includegraphics[width=\xwidth\linewidth]{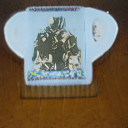} &
\includegraphics[width=\xwidth\linewidth]{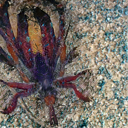}\\[\xem]
\includegraphics[width=\xwidth\linewidth]{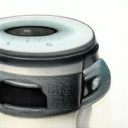} &
\includegraphics[width=\xwidth\linewidth]{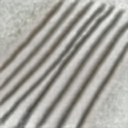} &
\includegraphics[width=\xwidth\linewidth]{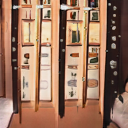} &
\includegraphics[width=\xwidth\linewidth]{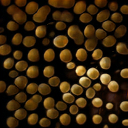} &
\includegraphics[width=\xwidth\linewidth]{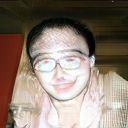} &
\includegraphics[width=\xwidth\linewidth]{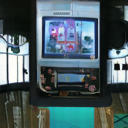}\\[\xem]
\includegraphics[width=\xwidth\linewidth]{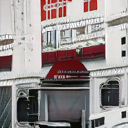} &
\includegraphics[width=\xwidth\linewidth]{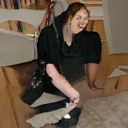} &
\includegraphics[width=\xwidth\linewidth]{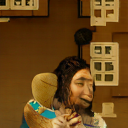} &
\includegraphics[width=\xwidth\linewidth]{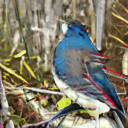} &
\includegraphics[width=\xwidth\linewidth]{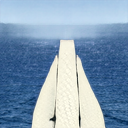} &
\includegraphics[width=\xwidth\linewidth]{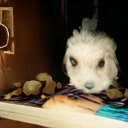}\\[\xem]
\includegraphics[width=\xwidth\linewidth]{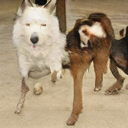} &
\includegraphics[width=\xwidth\linewidth]{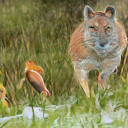} &
\includegraphics[width=\xwidth\linewidth]{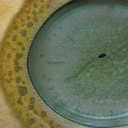} &
\includegraphics[width=\xwidth\linewidth]{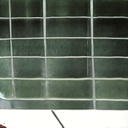} &
\includegraphics[width=\xwidth\linewidth]{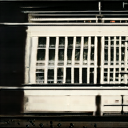} &
\includegraphics[width=\xwidth\linewidth]{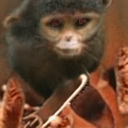}\\[\xem]
\end{tabular}
% \vspace{-0.35cm}
\caption{Unconditional generation results on ImageNet-1k $128\times 128$.}
% \vspace{-4pt}
\label{fig:more-gan-results}
\end{figure}

\clearpage
\bibliographystyle{splncs04}
\bibliography{egbib}
\end{document}